\documentclass{article} 
\usepackage{iclr2025_conference,times}

\usepackage{amsmath,amsfonts,bm}

\def\eqref#1{equation~\ref{#1}}

\def\1{\bm{1}}

\DeclareMathAlphabet{\mathsfit}{\encodingdefault}{\sfdefault}{m}{sl}
\SetMathAlphabet{\mathsfit}{bold}{\encodingdefault}{\sfdefault}{bx}{n}

\usepackage{makecell}  
\usepackage{booktabs} 
\usepackage{algorithm}
\usepackage{algpseudocode}
\usepackage{amsmath}
\usepackage{enumitem}
\usepackage{hyperref}
\usepackage{url}
\usepackage{tabularx}
\usepackage{booktabs}
\usepackage{cleveref}
\usepackage[frozencache,cachedir=.]{minted}
\usepackage{amssymb}
\usepackage[normalem]{ulem}

\definecolor{edgeblue}{RGB}{0, 0, 200}
\definecolor{edgegreen}{RGB}{0, 200, 0}
\definecolor{gptgreen}{RGB}{0, 166, 126}
\definecolor{scholarpurple}{RGB}{169, 1, 251}
\definecolor{bgcode}{rgb}{0.95,0.95,0.95}

\usepackage{xcolor}
\newcommand{\nbc}[3]{
 {\colorbox{#3}{\bfseries\sffamily\scriptsize\textcolor{white}{#1}}}
 {\textcolor{#3}{\sf\small\textit{#2}}}
 }

\definecolor{kscolor}{rgb}{0.9,0.1,0.1}
\definecolor{mscolor}{rgb}{0.1,0.1,0.9}
\definecolor{stcolor}{rgb}{0.1,0.9,0.1}

\usepackage{subcaption}
\usepackage{titling}

\pretitle{\centering\LARGE\bfseries}
\posttitle{\par}

\Crefname{section}{Sec.}{Sec.}

\title{Barbarians at the Gate:\\ How AI is Upending Systems Research}

\author{%
\begin{minipage}{\linewidth}\centering
\normalfont\normalsize
\textbf{Audrey Cheng\thanks{Equal contribution, ordered alphabetically.}, \,
Shu Liu,\textsuperscript{*} Melissa Pan\textsuperscript{*}, Zhifei Li, Bowen Wang, Alexander Krentsel,\\
Tian Xia, Mert Cemri, Jongseok Park, Shuo Yang, Jeff Chen, Lakshya Agrawal, \\ Aditya Desai, Jiarong Xing,
Koushik Sen, Matei Zaharia, Ion Stoica\\}
\small UC Berkeley
\end{minipage}
}

\newif\ifcomments
\newcommand{\eat}[1]{}
\ifcomments
    \providecommand{\shu}[1]{{\color{orange}{/* shu: #1 */}}}
    \providecommand{\melissa}[1]{{\color{magenta}{/* melissa: #1 */}}}
    \providecommand{\accheng}[1]{{\color{blue}{/* accheng: #1 */}}}
    \providecommand{\ion}[1]{{\color{cyan}{/* ion: #1 */}}}
    \providecommand{\jx}[1]{{\color{red}{/* xing: #1 */}}}
    \providecommand{\andyl}[1]{{\color{brown}{/* andyl: #1 */}}}
    \providecommand{\tian}[1]{{\color{teal}{/* tian: #1 */}}}
    \providecommand{\bowen}[1]{{\color{olive}{/* bowen: #1 */}}}
    \providecommand{\mert}[1]{{\color{olive}{/* mert: #1 */}}}
    \providecommand{\red}[1]{{\color{red}{/*#1 */}}}
    \providecommand{\ks}[1]{\nbc{KS}{#1}{kscolor}}
    \providecommand{\todo}[1]{
    {\colorbox{red}{\bfseries\sffamily\scriptsize\textcolor{white}{TODO}}}
     {\textcolor{red}{\sf\small\textit{#1}}}
    }
\else
    \providecommand{\shu}[1]{}
    \providecommand{\melissa}[1]{}    
    \providecommand{\accheng}[1]{}
    \providecommand{\ion}[1]{}
    \providecommand{\jx}[1]{}
    \providecommand{\andyl}[1]{}
    \providecommand{\tian}[1]{}
    \providecommand{\bowen}[1]{}
    \providecommand{\mert}[1]{}
    \providecommand{\red}[1]{}
    \providecommand{\ks}[1]{}
    \providecommand{\todo}[1]{}
\fi

\newcommand{\SYS}{}
\def\SYS/{ADRS}
\newcommand{\NumCase}{}
\def\NumCase/{eleven}
\newcommand{\numcase}{}
\def\numcase/{11}
\usepackage{tabularx} 
\usepackage{makecell} 
\usepackage{array}    
\usepackage{graphicx} 

\usepackage{tikz}
\usepackage{multirow}
\usepackage{tabularx} 
\usepackage{xcolor}
\usepackage{pifont}

\usepackage{enumitem}
\setitemize{noitemsep,topsep=0pt,parsep=0pt,partopsep=0pt}

\iclrfinalcopy 
\begin{document}

\maketitle

\begin{abstract}

Artificial Intelligence (AI) is starting to transform the research process as we know it by automating the discovery of new solutions. Given a task, the typical AI-driven approach is (i) to generate a set of diverse solutions, and then (ii) to verify these solutions and select one that solves the problem. Crucially, this approach assumes the existence of a reliable verifier, i.e., one that can accurately determine whether a solution solves the given problem. We argue that systems research, long focused on designing and evaluating new performance-oriented algorithms, is particularly well-suited for AI-driven solution discovery. This is because system performance problems naturally admit reliable verifiers: solutions are typically implemented in real systems or simulators, and verification reduces to running these software artifacts against predefined workloads and measuring performance. We term this approach as AI-Driven Research for Systems (ADRS), which iteratively generates, evaluates, and refines solutions. Using OpenEvolve, an existing open-source ADRS instance, we present case studies across diverse domains, including multi-region cloud scheduling, load balancing for Mixture-of-Experts inference, LLM-based SQL queries, and transaction scheduling. In multiple instances, ADRS discovers algorithms that outperform state-of-the-art human designs (e.g., achieving up to 5.0$\times$ runtime improvements or 26\% cost reductions). We distill best practices for guiding algorithm evolution, from prompt design to evaluator construction, for existing frameworks. We then discuss the broader implications for the systems community: as AI assumes a central role in algorithm design, we argue that human researchers will increasingly focus on problem formulation and strategic guidance. Our results highlight both the disruptive potential and the urgent need to adapt systems research practices in the age of AI.

\end{abstract}
\section{Introduction}

One of the most ambitious goals of artificial intelligence (AI) is to revolutionize scientific discovery by automating algorithm design, experiment execution, and even the research process itself. While the realization of this goal will likely be uneven—with certain domains being transformed earlier and more profoundly than others—AI-driven  approaches~\cite{alphaevolve,openevolve} have already reached a level of capability where they can meaningfully contribute to computer systems research. This raises fundamental questions about how we should conduct research.\ks{The previous sentence is pretty broad.  We are asking how we should do systems research with AI.}

A significant portion of systems research---spanning networking, databases, and distributed systems---is dedicated to enhancing performance. This is typically achieved through the meticulous, human-driven design of new algorithms for tasks such as routing~\cite{Boukerche-2011-ComputerNetworks, SirikaMahajan2016SurveyDynamicRouting, Suryanar23-Routing}, scheduling~\cite{wu2024can,cheng2024towards}, and resource management~\cite{Gao_2024,KhanPasrichaKim2020_PIMNMP_Survey}. Crucially, the novelty and efficacy of these algorithms are often the primary metrics for a publishable paper. \ks{Should we increase the scope of the domains?  Such as security, compiler optimization etc.?}

The central thesis of this paper is that a new class of AI-driven approaches, which we term AI-Driven Research for Systems (\SYS/), is beginning to show promising results in automated algorithm discovery, and will ultimately prompt a re-evaluation of the traditional role of systems researchers. To substantiate this claim, we follow previous work in presenting several case studies using existing \SYS/ frameworks (e.g., OpenEvolve~\cite{openevolve}) to generate algorithms that match or even exceed the performance of state-of-the-art, human-designed solutions for a range of computer systems research problems~\cite{wu2024can,deepseek-eplb}. 
For example, in a load-balancing problem for a Mixture-of-Experts (MoE) model, OpenEvolve discovered an algorithm to rebalance experts across GPUs that is 5.0$\times$ faster than the best-known baseline. In a job scheduling problem aimed at reducing costs by using spot instances across multiple cloud regions, OpenEvolve generated a solution that achieved roughly 30\% greater savings than an expert-developed baseline.
By using powerful LLMs such as GPT-4o, o3, and Gemini 2.5 Pro, most of our case studies produced solutions that matched or surpassed the state-of-the-art within a few hours, at a cost of only a few dollars to tens of dollars (see Table \ref{tab:project-summary}). We note that these results 
were obtained by different students without extensive ablation studies. Therefore, the reported results should be viewed as a lower bound on the capabilities of the \SYS/ frameworks. We expect stronger results as researchers gain more experience with these frameworks and as both the frameworks and their underlying models continue to improve.

One of the main reasons system performance problems are a particularly good fit for AI-driven research is that their solutions can be verified relatively easily. \ks{I would say that solutions could be verified automatically rather than easily.  Full verification of the entire system is intractable for most systems.  We are doing simulations to verify a handful of test cases, so it is far from verification. My point is that one should not accept the solutions just based on "verification"---instead, the researcher should manually validate the solutions.} The typical pattern for solving a problem using AI is to generate a diverse set of solutions and then verify which solutions actually solve the problem, if any~\cite{alphageometry,alphago}. Thus, verification accuracy is key to the success of an AI-driven approach in a given problem domain.
In general, verification is challenging. For example, it can be difficult to verify that an AI-generated program or an answer to a trivia question is correct. Fortunately, this is not the case for system performance problems. \ks{I am not sure if I agree with this.  If we are modifying code, we need full verification of functional equivalence. I think the verification problem remains intractable either way. We are simply conducting testing to identify and eliminate obvious mistakes.  It might be better to say that equivalence checking is simpler than full functional verification.  That is how we sold R2E.} In such cases, solutions—such as new algorithms and protocols—are typically integrated into artifacts like networks, databases, or operating systems. These solutions are then evaluated by running the system in which they are implemented against representative workloads and inspecting the resulting performance metrics. For instance, to evaluate a new routing protocol, we implement it in the routers of a network system and measure outputs like throughput, delays, loss rate for various workloads.
In another example, to evaluate a new CPU scheduler, we can implement it in an operating system, and then measure the response times for mixes of interactive and batch applications. 
Moreover, to avoid the high overhead of relying on real systems for evaluations, researchers often develop simulators that capture the salient features of real systems, enabling rapid and low-cost iteration. These simulators allow researchers to evaluate solutions quickly before deploying them in live systems, making systems research particularly well-suited for AI-driven exploration.

In the broader context of AI-driven research\ks{Please cite papers to concretize the context.}, our focus is deliberately narrow. Not only do we restrict our scope to the systems domain, but in this context, we focus only on the task of solution discovery, while largely ignoring other aspects in the research process, like problem formulation, literature survey, or paper writing. 
Narrowing the focus has several advantages. First, as discussed, it reduces the risk of hallucinations, since solution correctness can be grounded in empirical system performance. Second, it allows us to go deeper and develop stronger solutions within one domain, rather than spreading efforts across many. At the same time, if successful, the systems area is impactful enough to accelerate progress in other computer science areas by providing faster, more efficient infrastructures. 

\ks{Should we call this AI driven research because that phrase has a different meaning for frontier models.  It searches for related work and creates a survey.  Maybe AI-driven research Discovery is better.}
Finally, given this impending shift towards AI-driven research, we discuss its consequences for the systems community. As AI takes on the role of algorithm discovery and optimization, the emphasis for human researchers will likely pivot to problem formulation, high-level ideation, and strategic direction. In this new model, the researcher acts as an advisor to powerful AI research assistants: defining meaningful problems, proposing creative starting points, and distilling insights from generated solutions. This approach can create a powerful virtuous cycle: the same AI-driven methodologies can be applied to improve the AI systems themselves, leading to a compounding acceleration of the pace of discovery. 

This paper serves as a call to action for the systems community to proactively consider these changes, adapt our skills, and guide the co-evolution of human and AI in research. To facilitate this transition, we provide best practices for leveraging these technologies, outline their benefits, and highlight several key open problems. Ultimately, leveraging AI will enable researchers to dedicate their time to the most creative and fulfilling aspects of their work.

\ks{It might be useful here to give a summary of the experiments and the results. You may also want to list lessons learned.}

\section{Related Work}
\label{sec:related-work}  


\SYS/ builds on a long line of research that combines large-scale search with machine learning to tackle complex problems. 

\textbf{Pre-LLM AI for System Optimizations.}
Prior to LLMs, machine learning had already been widely applied to systems.
In databases, learned models were used for query optimization~\cite{Marcus_2019}, cardinality estimation~\cite{balsa, NeuroCard}, learned indexes~\cite{kraska2018caselearnedindexstructures}, and automated system tuning~\cite{vanaken2017automatic}.
Reinforcement learning (RL) and other learning-based techniques have advanced core networking problems, including congestion control~\cite{jay2018internet}, packet classification~\cite{liang2019neuralpacketclassification}, and topology modeling~\cite{Rusek_2020}. 
More broadly, RL has been applied in systems for scheduling over data processing workloads~\cite{Mao2019Decima}, physical device placement~\cite{Mirhoseini2017DeviceICML}, and video streaming~\cite{Du2020ServerDriven}.

\textbf{Automated discovery with learned approaches.} A series of advances in AI have demonstrated the power of automated discovery in increasingly complex scientific and computational domains. Early successes include AlphaGo~\cite{alphago} and AlphaZero~\cite{silver2018alphazero}, which demonstrated how search and RL can master games, and AlphaFold~\cite{alphafold}, which achieved breakthroughs in protein structure prediction. 
AlphaDev~\cite{alphadev} extended these ideas to discover efficient low-level algorithms, while AlphaChip~\cite{alphachip2024} uses RL to generate production chip layouts. Big Sleep~\cite{bigsleep} applies AI agents to detect security vulnerabilities in software. More recently, benchmarks, such as AlgoTune~\cite{press2025algotune}, have been developed to evaluate LLM abilities to optimize programs for different domains.

\textbf{LLM-based coding assistants.} LLMs have transformed code generation to accelerate the research process. Coding assistants like GitHub Copilot~\cite{GitHubCopilot}, Cursor~\cite{CursorAgent2024}, and Codex~\cite{OpenAICodex}, and Claude Code~\cite{ClaudeCode2025} help researchers rapidly prototype ideas, build simulators, and implement baselines. 
These tools enable high-level algorithmic ideas to be rapidly translated into working implementations. 
Recent work also explores using LLMs for performance-critical code~\cite{hong2025autocompllmdrivencodeoptimization}, such as GPU kernel code generation and optimization~\cite{ouyang2025kernelbench}, further illustrating their potential as building blocks for \SYS/.

\textbf{LLM-driven research.}
Beyond coding assistance, recent work leverages LLMs to automate larger parts of the research process. 
Frameworks such as AlphaEvolve~\cite{alphaevolve, nadga2025alphaevolve} and OpenEvolve~\cite{openevolve} use MAP elites algorithm and island-based population models to evolve and discover new algorithms; GEPA~\cite{agrawal2025gepa} employs reflective prompt evolution for better LLM generation; and LLM4AD~\cite{liu2024llm4ad}, which provides a unified platform for algorithm design. 
ShinkaEvolve~\cite{shinkaevolve} emphasizes sample-efficient evolution, EvoPrompt~\cite{evoprompt} focuses on using genetic algorithms to optimize prompts, MetaMuse~\cite{ma2025algorithmgenerationcreativeideation} presents a self-reflective framework for algorithm generation, and PolicySmith~\cite{dwivedula2025manmade} selects top-performing solutions in designing algorithms.

End-to-end research automation attempts are also emerging. MLGym~\cite{nathani2025mlgym} offers standardized benchmarks for AI research agents, while Code Researcher~\cite{singh2025code} explores LLM-based agents for large-scale software engineering tasks. Recent work on self-evolving AI agents systematically optimizes their internal components~\cite{fang2025comprehensive}. There is growing interest in applying these approaches to broader systems problems~\cite{liang2025nexthorizon,zhouneuripstalk}.

Our work focuses on automated algorithm discovery for the systems domain, where strong evaluators enable reliable verification -- a critical requirement for productive automation.

\section{Why AI-Driven Research for Systems?}
\label{why-systems}

In this paper, we advocate an AI-driven approach to systems performance problems.
While performance optimization is not the only focus of systems research, it remains a central one—a brief survey of top systems, networking, and database venues (NSDI, OSDI, SIGMOD, SOSP, and VLDB) shows that over one-third of published papers feature performance optimization algorithms as their core contribution. The main reason we believe an AI-driven approach is particularly well suited to such problems is that it is relatively straightforward to develop robust and cost-effective verification processes to evaluate candidate solutions.



First, it is easy to verify whether a given solution improves a system’s performance. Such solutions typically introduce new techniques or algorithms that are implemented directly within the systems they aim to optimize. Verification then reduces to running these systems under representative workloads and checking whether they outperform the baselines on the relevant performance metrics.

Second, the solutions for systems performance problems typically preserve the correctness of the original solution, or, at least, it is relatively easy to verify if they do. 
For instance, it is easy to verify whether a load-balancing algorithm schedules all assigned tasks or whether a network router forwards all the packets it receives.

Third, the portions of system code that usually must undergo evolution is often relatively small\shu{what does it mean by must undergo? I think we should make it clear that it's not just system code but also the "algorithm" or "solution" part of the code, otherwise can also evolve entire codebase or sth like that}, e.g., the core logic of a scheduler, load balancer, or resource allocator. This makes the generated code easier for humans to interpret which can further help with verifying its correctness. In all of our case studies, we were able to readily understand the generated solutions and identify their main ideas and techniques (see Section~\ref{sec:case_studies}). 
As these tools become more powerful, we expect their scope of modification to grow and extend across multiple components, e.g., when designing complex distributed protocols. Maintaining code interpretability in such cases is an important topic for future research.


Finally, systems researchers often use simulators to develop and evaluate solutions before deploying them in real systems. Simulator-based verification is relatively cheap. 
Thus, even if the search process is inefficient and produces more candidate solutions than is strictly necessary, their evaluation remains practical. 
For example, each of our case studies required only a few hours and cost no more than several tens of dollars. That said, building simulators for complex systems (e.g., operating systems, databases) that are not only inexpensive but also faithful is far from trivial, and it remains a topic for future research (see Section~\ref{sec:better-evaluators}).



\section{Using AI to Accelerate Systems Research}
This section provides an overview of the systems research process and then introduces the AI-Driven Research for Systems (\SYS/) approach which accelerates this process through automatic solution discovery and evaluation.


\subsection{Systems Performance Research Process}
The typical research process employed by systems researchers can take many weeks or months. Broadly speaking, it consists of five stages (Figure~\ref{fig:research-process}): 

\begin{figure}[h!]
  \centering
  \vspace{-0.7em}
    \includegraphics[width=0.95\textwidth]{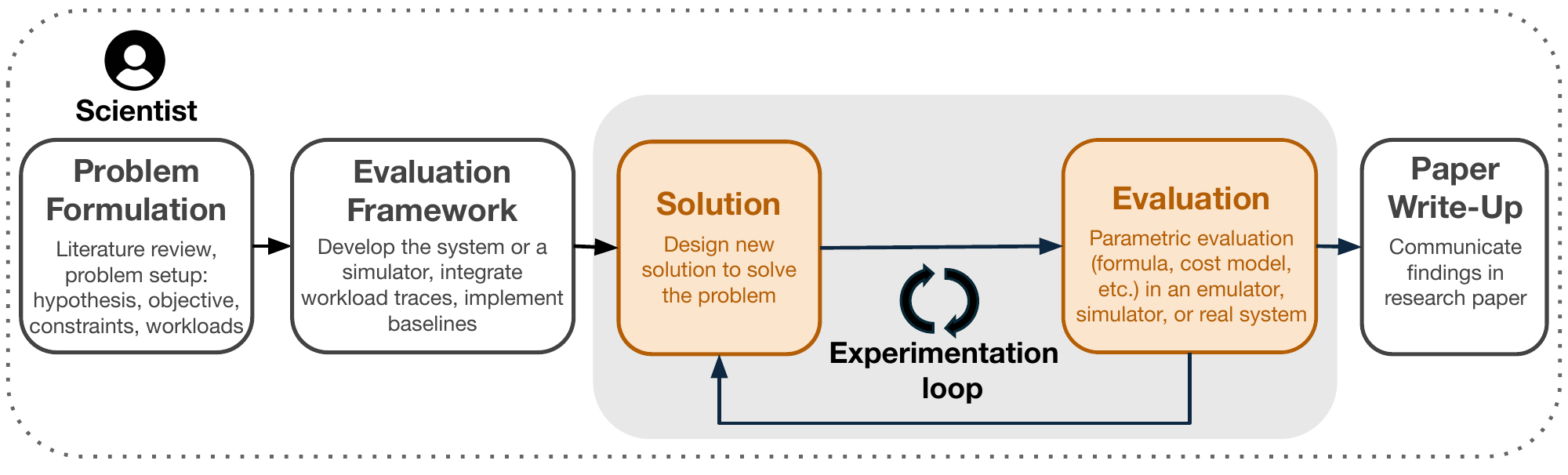}
  \caption{The five stages of the systems research process. In this paper, we show how AI can automate the \textbf{Solution} and \textbf{Evaluation} stages (grey area).}
  \vspace{-0.5em}
  \label{fig:research-process}
\end{figure}

\begin{itemize} 

\item \textbf{Problem Formulation:} Define the problem to solve, such as improving system throughput or latency. The entire research process is organized around solving this problem and communicating the results. 

\item \textbf{Evaluation Framework:} Develop a framework to implement and evaluate potential solutions. This framework can be the system itself or a simulator that approximates the system's behavior. Even when a system is available, a simulator may be built to accelerate development. In addition, researchers collect or use existing workloads (traces) or benchmarks to drive the evaluation. 

\item \textbf{Solution:} Develop a solution (e.g., algorithm) to the problem, such as a new scheduling, resource allocation, or search algorithm. 

\item \textbf{Evaluation:} Implement the solution in the simulator or system, evaluate its performance using the selected workloads, and compare the results against the baseline(s). If the solution does not show improvements, researchers return to the \textbf{Solution} stage to refine the approach or develop alternatives. 


\item \textbf{Paper Write-Up:} Once a solution achieves the desired results, document the findings for publication. 

\end{itemize}

\begin{figure}[h!]
  \centering
  \vspace{-0.7em}
    \includegraphics[width=0.5\textwidth]{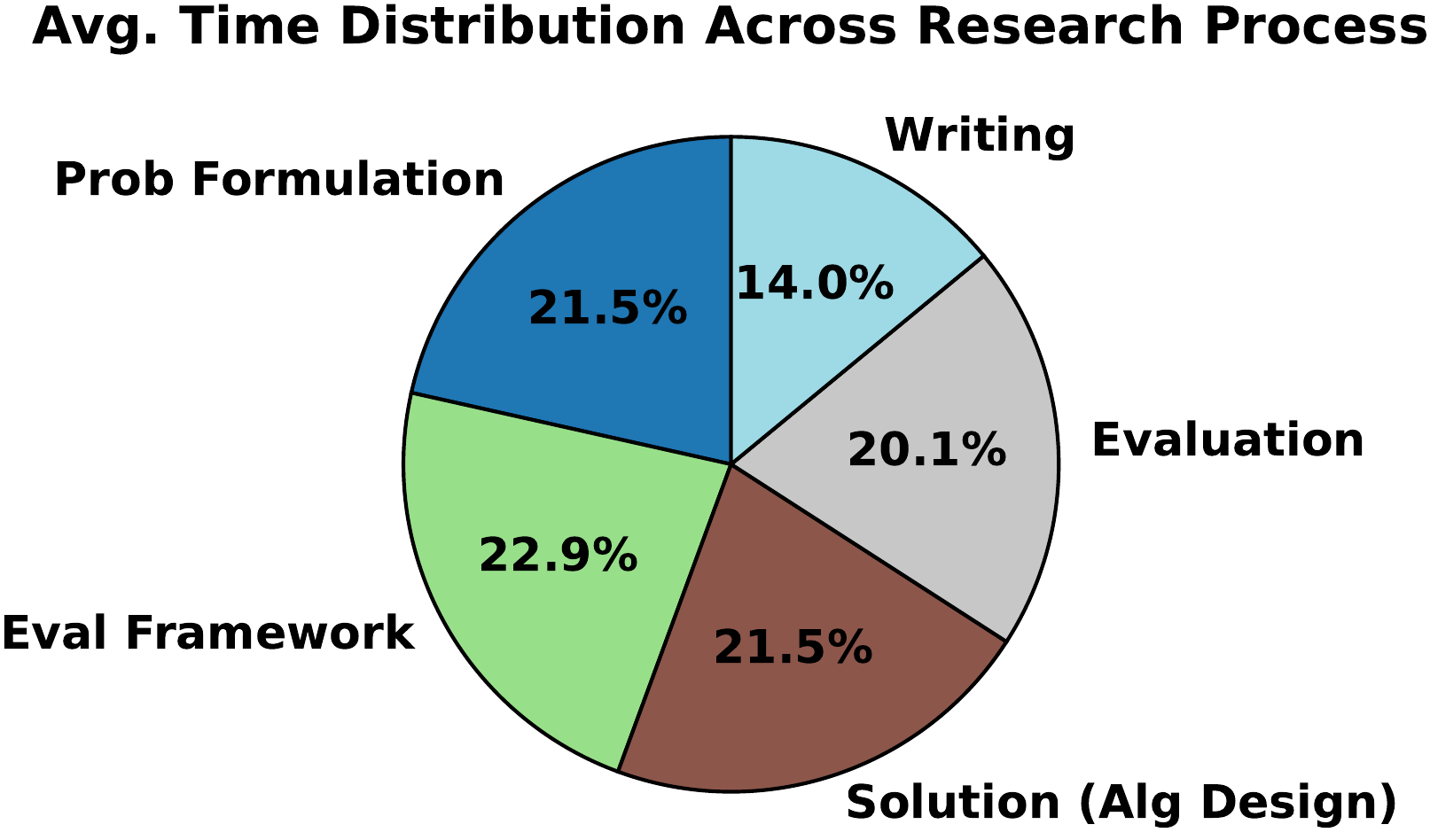}
  \caption{Time spent in various stages of the systems research process in systems, based on a survey of 31 PhD students. \textit{Algorithm Design} (21.5\%) and \textit{Evaluation} (20.1\%) together account for over 40\% of total effort, highlighting a significant opportunity for leveraging AI to accelerate this process.} 
  \vspace{-0.2em}
  \label{fig:research-process-survey}
\end{figure}

Figure~\ref{fig:research-process-survey} shows the approximate time spent in each of these stages, based on survey data from over 30 systems graduate students from a US university.

\subsection{AI-Driven Research for Systems (\SYS/)}

\begin{figure}[h!]
  \centering
  \vspace{-0.7em}
    \includegraphics[width=0.9\textwidth]{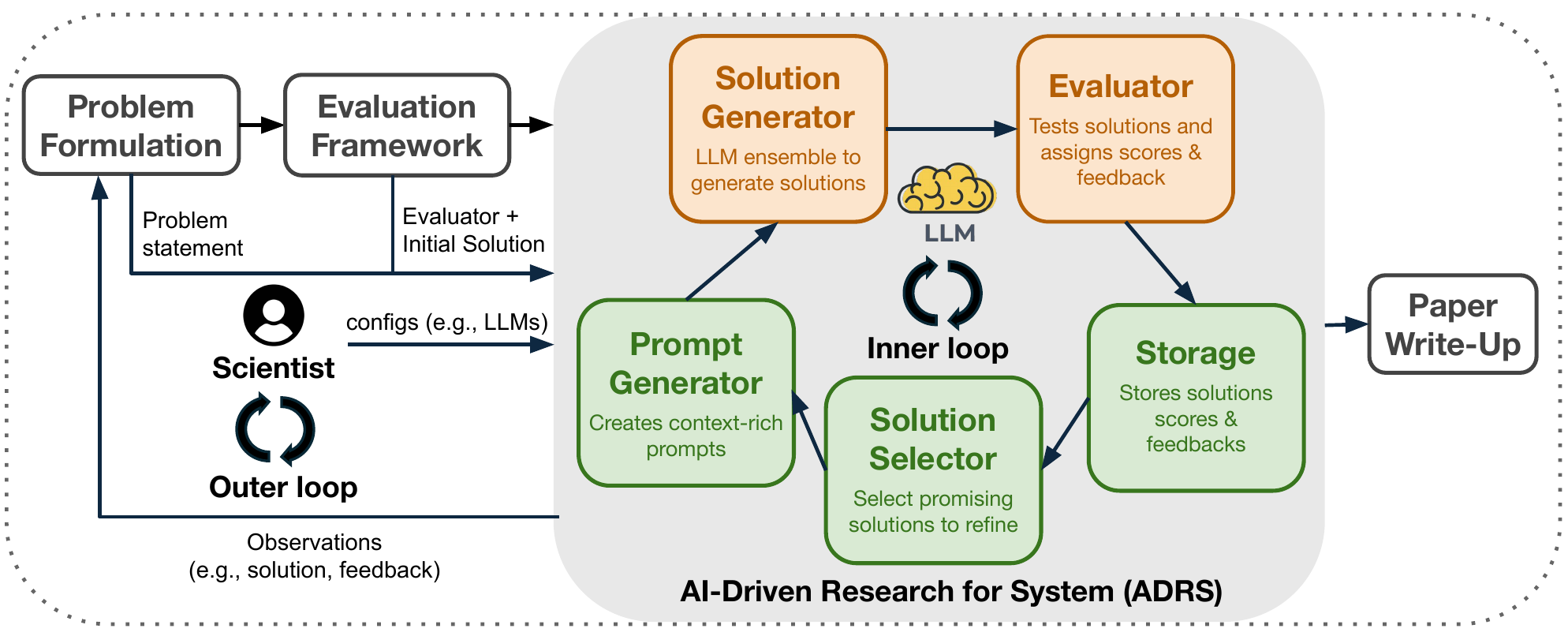}
  \caption{The AI-driven Research for Systems (\SYS/) architecture shown in the context of the systems research process (see in Figure~\ref{fig:research-process}). \SYS/ (grey area) automates the \textbf{Solution} and \textbf{Evaluation} stages.}
  \vspace{-0.5em}
  \label{fig:evolve-sys}
\end{figure}

\eat{
\begin{figure}[h!]
  \centering
  \vspace{-0.7em}
  \begin{subfigure}[t]{0.48\textwidth}
    \centering
    \includegraphics[width=\textwidth]{sections/figures/evolve-sys.pdf}
    \caption{AI-Driven Research for Systems architecture.} 
    \label{fig:evolve-sys}
  \end{subfigure}%
  \hfill
  \begin{subfigure}[t]{0.48\textwidth}
    \centering
    \includegraphics[width=\textwidth]{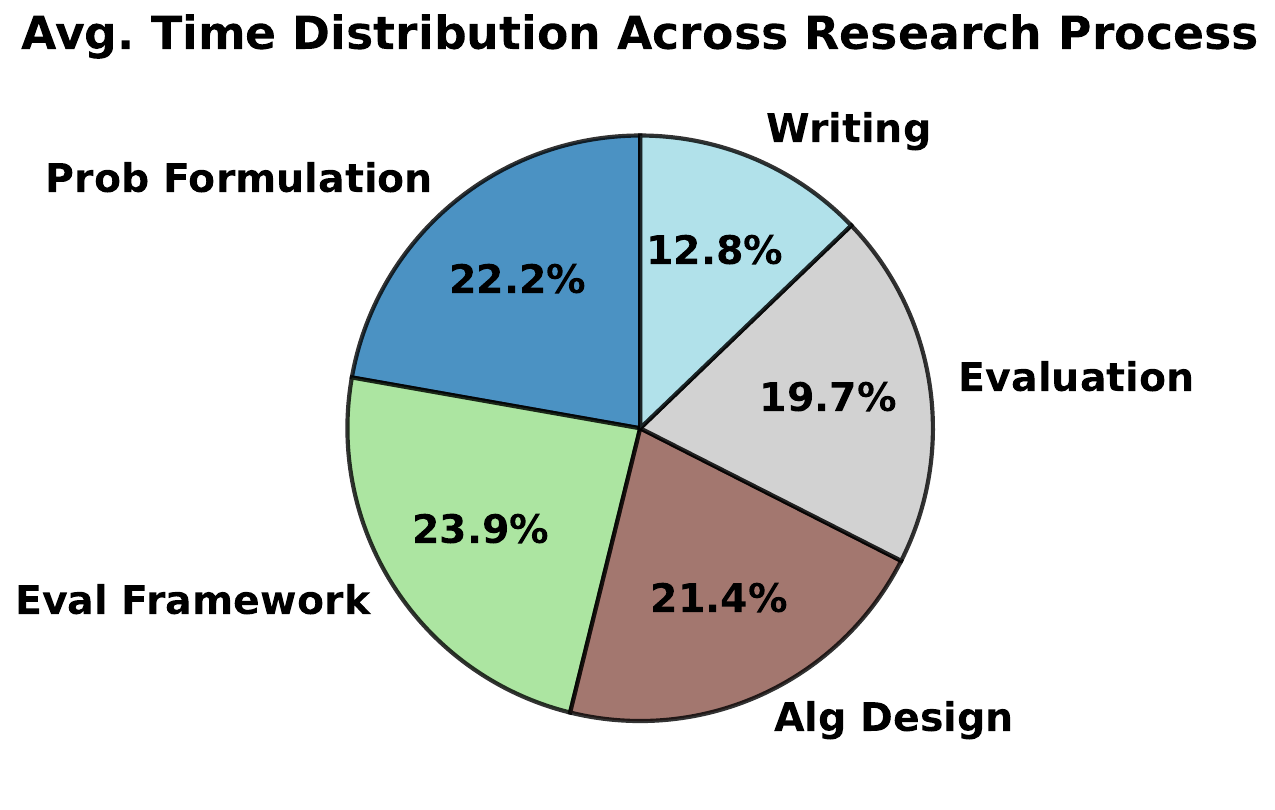}
    \caption{Research Process Survey.} 
    \label{fig:alphaevolve}
  \end{subfigure}
  \vspace{-0.5em}
  \caption{Comparison of system architecture (left) and survey process (right).}
\end{figure}
}


As Large Language Models (LLMs) have progressed from simple text completion to sophisticated reasoning and tool use, new architectures have been developed to enhance the reliability and scope of tasks to which they can be applied. In this paper, we focus on how LLMs can be used to design, implement, and evaluate new solutions (e.g., algorithms) to solve systems research problems. We call this approach the AI-Driven Research for Systems (\SYS/) and depict it in Figure~\ref{fig:evolve-sys}. Basically, \SYS/ implements the two iterative stages of the systems research process shown in Figure~\ref{fig:research-process}: \textbf{Solution} and \textbf{Evaluation}. Together, these two stages account for about 40\% of the time spent (based on a survey of over 30 systems researchers, Figure~\ref{fig:research-process-survey}) in producing new results. 

At its core, \SYS/ implements a loop that, in each iteration, creates or refines prompts for LLMs to generate new solutions or improve existing ones, and then evaluates these solutions in either a real system or a simulator. This iterative process continues until a desired solution is discovered, a resource budget is exhausted, or a researcher decides to stop it. \SYS/ consists of five components. 

\begin{itemize} 

\item \emph{Prompt Generator}: Creates the prompt used to generate the solution. This prompt consists of the problem statement and context provided by the researcher, which might include system or simulator code. The prompt may also include previous solutions and their evaluations (provided by the \emph{Solution Selector}) to further refine the prompt. 


\item \emph{Solution Generator}: Feeds the prompt from \emph{Prompt Generator} to one or more LLMs to generate a new solution or refine an existing one. This is typically done by directly updating the code in the simulator or the real system.

\item \emph{Evaluator}: Takes the solution from \emph{Solution Generator} and runs it against a predefined set of workloads (traces). It then scores the solution based on the run's performance and can use an LLM to provide qualitative feedback. If the score is high enough, the loop terminates.

\item \emph{Storage}: Stores the solutions, their outputs, scores, and the feedback provided by the \emph{Evaluator} component.

\item \emph{Solution Selector}: Chooses a subset of solutions from \emph{Store} and provides them to \emph{Prompt Generator} to refine the prompt to generate new and improved solutions.

\end{itemize}

Together, these components form an internal automated feedback loop that enables \SYS/ to iteratively refine solutions for a given problem. This can be paired with an outer feedback loop in which a human can observe the generated solutions and provide high-level guidance that is incorporated into future prompts. 

In most cases, \SYS/ relies on a simulator rather than the real system for solution implementation and evaluation. There are two main reasons for this. First, the codebase of the real system is often too large to fit within the context window of current LLMs. 
Second, running evaluations in a simulator can be orders of magnitude faster than on the real system, significantly accelerating the evolution process. 

As mentioned earlier, \SYS/ implements the \textbf{Solution} and \textbf{Evaluation} stages of the research process depicted in Figure~\ref{fig:research-process}: the \textbf{Solution} stage is implemented by the \emph{Prompt Generator} and \emph{Solution Generator} components, while the \textbf{Evaluation} stage is implemented by the \emph{Evaluator} component. As discussed, the \emph{Store} and \emph{Solution Selector} are auxiliary components that support the \emph{Prompt Generator}. Regarding the other stages in the research process, \SYS/ does not have any impact: it makes it neither harder nor easier for researchers to decide what problem to work on, building the evaluation framework, document their findings and write the paper. As a result, \SYS/ advances the Pareto frontier of the research process.

\subsection{\SYS/ Examples}
\label{sec:adrs-examples}

\SYS/ is not entirely a new concept. It aims to capture the architecture of recently developed systems designed for similar problems that allow for strong verification. Next, we provide several examples of such systems.

\textbf{AlphaEvolve and OpenEvolve.} 
AlphaEvolve~\cite{alphaevolve} is a system developed by Google DeepMind that uses artificial intelligence to automatically discover and design novel, high-performance computer science algorithms. 
Similarly to \SYS/, AlphaEvolve takes as input a user-provided problem definition, an initial reference solution, and a programmatic evaluation function that quantitatively scores any proposed solution. 
AlphaEvolve uses an evolution algorithm to facilitate the inner feedback loop, and it does not include an outer loop to take in human feedback during the evolution process (though such loop can be easily added).
The evolution algorithm uses a combination of the Multi-dimensional Archive of Phenotypic (MAP) elites algorithm~\cite{mouret_clune_2015_mapelites}  and island-based population models~\cite{tanese1989distributed} that the database not only contains the best-performing programs but also a wide variety of high-quality solutions with different attributes.

OpenEvolve~\cite{openevolve} is an open-source implementation of the main concepts presented in AlphaEvolve. OpenEvolve provides a controller that orchestrates an asynchronous pipeline between an LLM ensemble for code generation, a pool of evaluators for scoring, and a program database for storing and sampling candidate solutions. While it contains the core key features, it does not implement all functionality described in AlphaEvolve (e.g., meta prompt evolution) and provides some additional functionality (e.g., evolution tree visualization). Due to its availability as open source and flexibility, this is the main framework we have used in our study so far. 

\textbf{GEPA.} GEPA (Genetic-Pareto)~\cite{agrawal2025gepa} is another recent \SYS/-like framework that uses a different evolution process and reflection with natural language feedback to improve prompts. The evolution algorithm automates the exploration–exploitation tradeoff by iteratively mutating prompts with reflective updates, merging candidates, and filtering through Pareto frontiers to preserve diverse high-performing solutions. 


\textbf{LLM4AD.} LLM4AD~\cite{liu2024llm4ad} is a general-purpose platform that integrates search, LLM generation, and evaluation in a unifed framework. It provides modular components from problem specification, solution generation, evaluation, and integrates a broad set of search strategies with a standard interface, including evolutionary algorithms (e.g., NSGA-II, MOEA/D), neighborhood search, and random sampling. It also offers unified evaluation sandboxes, logging and profiling tools, and support for over 20 tasks. 
LLM4AD can be viewed as an end-to-end testbed that enables researchers to benchmark \SYS/ pipelines across diverse domains.



\textbf{Coding Assistants.} 
Coding assistants such as Cursor or Windsurf can also be seen as \SYS/ instances. These assistants have context over entire codebases, enabling them to perform complex code modifications. These coding assistants can be instructed to explore the solution space to evolve algorithms, use tools to verify them, and can orchestrate the entire feedback loop via natural language prompts. In addition, these tools facilitate a human-driven outer feedback loop by providing an easy way for the researcher to interactively provide guidance in the search.
\section{Evaluation and Case Studies}
\label{sec:case_studies}

\begin{table*}[t]
\centering
\scriptsize
\setlength{\tabcolsep}{4pt}
\renewcommand{\arraystretch}{1.2}

\begin{tabularx}{\textwidth}{@{}p{2.8cm} p{3.3cm} X p{2.2cm}@{}}
\toprule
\textbf{Task \& SOTA Publication} & \textbf{Objective} & \textbf{Result vs. SOTA / Baseline} & \textbf{Time / Cost} \\
\midrule

Telemetry Repair\newline \cite{krentsel2024case} \newline [\textit{HotNets `24}] &
Repair buggy network telemetry counters and calibration. &
+9\% better counter repair score, +30\% higher confidence calibration score than published solution. &
8h (300 iters), \newline $\leq\$10$ \\
\midrule

Adaptive Weight \newline Compression \newline [In-Progress Work] &
Assign bitrate per column to minimize bits/elem while preserving accuracy. &
Similar bits/elem, 14.2\% worse PPL. &
12h (200 iters), \newline $\leq\$20$ \\
\midrule

Cloudcast\newline \cite{wooders2024cloudcast} \newline [\textit{NSDI `24}] &
Optimize multi-cloud data transfer cost. &
Matches SOTA cost. & 
1h (100 iters), \newline $\leq\$5$ \\
\midrule

Expert Parallelism \newline Load Balancer \newline [In-Progress Work] &
Balance expert-parallel load across GPUs. &
Same balanceness, 2$\times$ faster runtime vs.\ internal implementation. &
5h (300 iters), \newline $\leq\$10$ \\
\midrule

Global Model Placement\newline \cite{yu2025prism} \newline[arXiv] &
Optimize cost for model-to-GPU placement. &
18.5\% cheaper than published solution. &
40m (70 iters), \newline $\leq\$5$ \\
\midrule

LLM-SQL\newline \cite{liu2024optimizing} \newline [\textit{MLSys `25}] &
Reorder tabular data to improve prefix hit rate. &
Comparable hit rate, 3.9$\times$ faster runtime. &
1h (100 iters), \newline $\leq\$7$ \\
\midrule

Transaction Scheduling\newline \cite{cheng2024towards}\newline [\textit{VLDB `24}] &
Minimize makespan in transaction scheduling. &
20\% better than greedy (offline). &
$<$2h (100 iters), \newline $\leq\$20$ \\
\midrule

Can’t Be Late \newline \cite{wu2024can} \newline [\textit{NSDI `24}]  &
Schedule deadline-driven jobs on single-region spot instances. &
Up to 16\% (average 7\%) higher cost savings vs.\ SOTA. &
5h (400 iters), \newline $\leq\$20$ \\
\midrule

Can’t Be Late\newline Multi-Region Extension \newline [In-Progress Work] &
Schedule deadline-driven jobs on multi-region spot instances. &
26\% lower cost vs.\ single-region baseline. &
1h (100 iters), \newline $\leq\$5$ \\
\midrule

Sparse Attention Design \newline \melissa{missing SOTA citation, added HashAttetion based on what i know, but should confirm with aditya} \cite{hashattention} \newline [\textit{NeurIPS `25}] &
Balance attention sparsity and accuracy. &
7\% average error and density improvement vs.\ SOTA. &
4h (100 iterations), \newline $\leq\$15$ \\
\midrule

Multi-Agent System \newline Optimization \newline [\textit{ICLR `24}] &
Improve multi-agent collaboration using MAST taxonomy and rubric-based feedback. &
7\% improvement on ProgramDev. &
$<$2h (100 iters), \newline  $\leq\$15$ \\

\bottomrule
\end{tabularx}

\caption[Summary of project tasks]{Summary of project task objectives and it's corresponding SOTA publication, performance improvements relative to SOTA and baseline solutions, and overall time/cost efficiency. 
Most tasks achieve near-SOTA performance within hours at modest cost, demonstrating the practicality of the ADRS approach.\protect\footnotemark \shu{double check the sparse attention results}}
\label{tab:project-summary}
\end{table*}

\footnotetext{Reported “time” reflects the automated solution-iteration phase (model–evaluator loop). It excludes researcher effort related to problem specification (e.g., refining prompts, evaluators, or experimental setup), which typically ranges from a few hours to a few days—still orders of magnitude less than the time required to manually develop a solution. Because this effort is difficult to measure consistently, we do not report it here.}

\begin{table}[h]
\centering
\small
\begin{tabular}{p{0.25\linewidth}p{0.68\linewidth}}
\toprule
\textbf{Dimension} & \textbf{Key Components} \\
\midrule
\textbf{Prompt Generator (Problem Setup)} 
& \textit{Problem description}: problem domains -- computer systems (networking, distributed systems, databases, MLSys, etc.); problem description -- performance optimization, e.g., find the most cost-effective transfer graph \\
&\\
& \textit{Optimization objective}: e.g., latency, throughput, cost, algorithm runtime \\
& \textit{Constraints}: e.g., latency SLOs \\
\midrule
\textbf{Solution Generator} 
& \textit{LLM type}: what model used for solution generation, this includes reasoning vs.\ non-reasoning, tool-use vs.\ non-tool-use, LLM ensemble\\
&\\
& \textit{Number of iterations}: number of rounds to iterate the solution\\
\midrule
\textbf{Evaluator}
& \textit{Environment and test data}: testbed environment such as CPU simulator, database, GPU environment; test data and traces \\
&\\ 
& \textit{Initial Program}: the initial program to start with, use public source (GitHub/paper) if available, otherwise use simple algorithm \\

& \textit{Additional Baselines}: additional baselines for comparison if any \\
&\\
& \textit{Evaluator Feedback}: execution score, more advanced if any (e.g., error messages, human- or agent-in-the-loop feedback) \\

\midrule
\textbf{Solution Selector} 
& \textit{Selection algorithm}: e.g., greedy, random, or island algorithm \\
\specialrule{1pt}{0.4em}{0.4em} 
\textbf{How We Analyze Result} 
& \textit{Performance}: compare evolved result vs.\ initial program and SOTA algorithm \\
& \textit{Cost-benefit}: compute budget, LLM calls, simulation time (survey: was the quality gain worth the cost?)\\
& \textit{Other metrics}: robustness, generalization \\
\midrule
\textbf{How We Analyze Evolution Process} 
& \textit{Search trajectory}: from initial program to checkpoints to final outputs (examples of key transitions, what features are added) \\
&\\
& \textit{Common patterns}: where models get stuck, recurring failures \\
& \\
& \textit{Feedback granularity and utility}: scores, constraint violations, trace-level logs; which kinds of feedback resolve which failure patterns\\
\bottomrule
\end{tabular}
\caption{Expanded schema for problem formulation and evaluation of AI-driven algorithm discovery. Each row corresponds to an element in the setup.}
\label{tab:expanded-schema}
\end{table}

To evaluate the \SYS/ approach, we investigate \numcase/ system tasks across several sub-domains, including networking, databases, and core systems. We summarize our findings in Table~\ref{tab:project-summary}. Each case study follows a common schema that captures the problem setup, environment, evolutionary process, model usage, and final outcome. We show this schema in Table~\ref{tab:expanded-schema}.

Next, we present four representative case studies to highlight insights that we hope will be useful to other researchers. These insights cover both the limitations of current frameworks and best practices for using them, which are detailed further in Section~\ref{sec:best-practices}. We used OpenEvolve as the primary \SYS/ for our case studies. 

The four case studies cover distributed systems, databases, and LLM systems.

\begin{itemize}
\item \emph{Optimizing Spot Instance Savings under Deadlines:} Given a job with a deadline, the solution aims to maximize the use of cheaper spot instances in a public cloud without violating the deadline. \SYS/ improves the SOTA result by up to 16\% for a single region\shu{can we report this "up to" number instead?} and achieves 48\% improvements over a strong baseline in a multiple-region setting. 

\item \emph{Optimizing Expert Placement in MoE Inference:} The solution seeks to balance the load across GPUs by mapping the expert replicas across GPUs. \SYS/ provides a fivefold improvement in the time it takes to rebalance experts compared with the best-known proprietary implementation.

\item \emph{Optimizing LLM Inference for SQL Queries:} The solution to this problem reorders rows and columns in a table to maximize the hit rate in a KV cache when performing LLM inference. \SYS/ achieves a similar hit rate to SOTA, while reducing the running time of the reordering algorithm by 3$\times$.

\item \emph{Optimizing Transaction Scheduling:} The solution aims to reorder transactions to minimize conflicts and hence improve the makespan and throughput. \SYS/ ``rediscovers'' the SOTA solution for the online case and improves a strong baseline by 34\% for the offline case, for which we are not aware of any published solution.
\end{itemize}

Before proceeding, we make one important point. The case studies we present here were carried out by different students in parallel during the summer of 2025. As a result, they use (sometimes widely) different configuration parameters. This makes direct comparison difficult and highlights the need for systematic ablation studies to identify which configurations work best for different problems—an important direction for future work (see Section~\ref{sec:future-work-framework}). Consequently, the case studies presented here should be viewed as a lower bound on the capabilities of the existing \SYS/ framework. As we gain deeper understanding of how to use these frameworks effectively and as the frameworks themselves evolve, we expect to see even more impressive results.
\mert{It is nice to mention these results as ``lower bound", but in addition one can frame this as a practical robustness of \SYS/: even with out of the box and not so optimized configs, we can get good results across different applications.}



\subsection{Case Study \# 1: Optimizing Spot Instance Savings under Deadlines}
\label{sec:spot-instances}

Our first case study focus on reducing the cost of deadline-driven jobs by exploiting cheaper but unreliable spot instances in the cloud. Spot instances are typically 60\% to 90\% cheaper than on-demand instances, but they are not always available and can be preempted at any time. Each preemption incurs a changeover delay, representing the setup time on a new instance. The challenge is to minimize the cost by using spot instances as much as possible without violating job deadlines.

We use OpenEvolve to evolve algorithms for this setting. It discovers an algorithm that improves the cost savings of the best published solution by an average of 7\%, with per-workload improvements of up to 16.7\%. We also use OpenEvolve to develop an algorithm on an expanded multi-region setting, where no prior policy has been published. In this setting, OpenEvolve discovers an algorithm that outperforms a hand-tuned baseline by 26\%.

\subsubsection{Single Region: Can't Be Late [NSDI `24]}
\label{sec:cant-be-late}

\begin{figure}[h!]
\caption{Side-by-side comparison of the initial Uniform Progress policy and the evolved adaptive strategy. Key innovations in the evolved policy are highlighted.}
\label{fig:cant_be_late_evolution_comparison}
\centering

\begin{subfigure}{0.45\textwidth}
    \begin{minted}[
        frame=lines,
        fontsize=\tiny,
        linenos,
        breaklines=true
    ]{python}
def UniformProgress(has_spot, state, env, task):
    # Calculate uniform progress rate
    progress_rate = task.duration / task.deadline
    expected = env.elapsed_time * progress_rate
    actual = task.progress_made

    # Simple decision: behind schedule?
    if actual < expected:
        # Must use any available resource
        if has_spot:
            return ClusterType.SPOT
        else:
            return ClusterType.ON_DEMAND

    # Hysteresis for on-demand
    if state == ClusterType.ON_DEMAND:
        buffer = expected + 2 * task.overhead
        if actual < buffer:
            return ClusterType.ON_DEMAND

    # Default: use spot if available
    if has_spot:
        return ClusterType.SPOT
    else:
        return ClusterType.NONE  # Wait
      \end{minted}
    \caption{Before: Uniform Progress Policy}
    \label{fig:uniform-progress-code}
\end{subfigure}
\hfill 
%
\begin{subfigure}{0.45\textwidth}
    \begin{minted}[
        frame=lines,
        fontsize=\tiny,
        linenos,
        breaklines=true,
        highlightlines={3-8,24-26,28-29,31-32,39-40},
        highlightcolor=blue!10
    ]{python}
def AdaptiveStrategy(has_spot, state, env, task):
    # Track recent spot availability patterns
    self.window.append(has_spot)
    alpha = window_avg(self.window)
    streak = longest_run(self.window)
    tail = trailing_run(self.window)

    params = get_params(classify_situation(alpha, streak, tail))

    # Compare remaining time to safety margin
    need = ticks_needed(task, env)
    slack = ticks_remaining(env)
    safety = safety_margin(task, params)

    if need >= slack:
        lock_on_demand()
        return ClusterType.ON_DEMAND

    if is_locked():
        if can_unlock(has_spot, tail, params):
            unlock()
        else:
            return ClusterType.ON_DEMAND
    elif slack <= need + params.lock_margin:
        lock_on_demand()
        return ClusterType.ON_DEMAND

    if rebuilding_buffer() and not buffer_recovered(slack, safety, params):
        return ClusterType.ON_DEMAND

    if safety >= slack:
        return ClusterType.SPOT if has_spot else ClusterType.ON_DEMAND

    if state == ClusterType.ON_DEMAND:
        if safe_to_switch(has_spot, streak, tail, params):
            return ClusterType.SPOT
        return ClusterType.ON_DEMAND

    if has_spot and looks_stable(streak, tail, params):
        return ClusterType.SPOT
    elif slack > safety + params.wait_margin:
        return ClusterType.NONE  # Wait for stable spot
    else:
        start_rebuilding(params.dwell)
        return ClusterType.ON_DEMAND
    \end{minted}
    \caption{After: Evolved Adaptive Policy}
    \label{fig:adaptive-evolved-code}
\end{subfigure}

\end{figure}

First, we discuss optimizing cost savings for a single region. This problem was explored in a \textit{NSDI `24} outstanding paper~\cite{wu2024can}, which introduced the current state-of-the-art policy.

\textbf{Problem setup.} 
The task is to minimize the cost of running a deadline-aware job on a single node by using spot instances in one cloud region, while ensuring the job still meets its deadline. 

\textit{Objective and constraints.} We evaluate the average cost savings across a range of workload traces. We require that all deadlines be met for an algorithm to be considered valid.

\textbf{Solution generator and selector.} We run OpenEvolve's default island evolution with 4 islands for 400 iterations, which takes 5 hours total and costs less than \$20. We use a two-model ensemble: 20\% GPT-5 for reasoning-driven exploration and 80\% Gemini 2.5 Pro to enhance diversity.

\textbf{Evaluator.} We use the simulator from the published paper and use configurations covering different job fractions, changeover delays, regions, and accelerator types. 
For each configuration, we sample 30\% of the traces as a feedback subset used during OpenEvolve search to reduce overfitting to specific traces.
We report final results on the full evaluation set.

\textit{Initial program and baselines.}  
The initial program is the \emph{greedy policy} from~\cite{wu2024can}, which uses spot instances until preemption risks missing a deadline. 
We compare against the \emph{Uniform Progress} algorithm, the paper's state-of-the-art solution, which tracks expected progress and switches between spot and on-demand instances based on whether it is ahead or behind schedule.


\textit{Evaluator Feedback.} The evaluator checks syntax and interface compliance and tests valid solutions on sampled traces. 
It reports average cost savings over the Uniform Progress baseline and per-configuration statistics (mean, deviation, count). Trace features such as availability and average spot duration are also included to provide richer context.

\textbf{OpenEvolve results.} 
The best policy is developed in iteration 389, achieving 7\% higher average cost savings than Uniform Progress (and 23.8\% over the greedy policy) while meeting all deadlines. 
Per-trace improvements reach up to 16.7\% compared to Uniform Progress.

As shown in Figure~\ref{fig:adaptive-evolved-code}, the evolved policy fundamentally differs from Uniform Progress shown in Figure~\ref{fig:uniform-progress-code}. 
While Uniform Progress follows a fixed formula to maintain steady progress, the evolved policy adaptively learns from recent spot availability patterns based on a sliding window (lines 3). 
It classifies situations as stable, moderate, or unstable and adjusts action accordingly (lines 4-8). 
When spots are stable, it risks more to save cost; when unstable, it becomes conservative.
The policy compares remaining slack time to safety margins that expand or shrink based on recent spot stability (lines 24-26). 
When slack becomes tight or spots look unreliable, it temporarily switches to on-demand instances to rebuild the slack buffer, preventing last-minute scrambles (lines 28-29). 
When slack recovers and patterns look favorable again, it returns to using spots. 
The safety margins scale dynamically with the cost of recent preemption (lines 31-32), creating larger buffers when restart costs are high.

The main limitation of Uniform Progress is its inflexibility when behind schedule: it must use every available spot instance, regardless of how short-lived it might be. This leads to a “changeover trap,” where frequent, brief spot allocations cause repeated switches with little real progress. The evolved policy avoids this by making spot use selective (lines 39–40): it waits when spots appear unstable and sufficient slack remains, rather than wasting effort on likely short-lived opportunities. This selective waiting lets it skip unreliable spots while still exploiting stable ones when conditions improve.


\textbf{Evolution process.} The search runs for 400 iterations, gradually refining when to use spot instances.
In early iterations (1–90), the policy learns to track recent spot availability using a sliding window and recognize when spots are plentiful versus scarce. 
By iteration 180, it introduces adaptive safety margins adjustment based on spot stability.
Around iteration 240, it adds finer-grained situation detection, using hysteretic transitions between stable and unstable regimes to avoid frequent switching.
By iteration 350, it learns that safety margins should adapt dynamically through chance-constrained risk lines, expanding when recent preemption are frequent and shrinking when spots have been stable. 
The final strategy (iteration 389) integrates all these mechanisms with lean parameter tuning and introduces selective waiting:
skipping unreliable spots and waiting for stable ones.


\subsubsection{Multi-Region Can't Be Late}
\label{sec:cant-be-late-multiple-regions}

\textbf{Problem setup.} The original Uniform Progress assumes one region with uniform spot prices. In practice, spot prices and availability differ across regions (\cite{10.1145/3689031.3717459}), so a policy must decide when to switch spot and on-demand, which region to use, and when to migrate jobs. We use OpenEvolve to explore this multi-region space and derive an better policy.

\textit{Objective and constraints.} We evaluate the total cost in a multi-region setup, accounting for spot and on-demand instances and migration costs. A policy is valid only if all job deadlines are met.


\textbf{Evaluator and other config.}  
For the multi-region setting, we extend the Uniform Progress simulator and evaluate cost savings on 106 traces. As no baseline exists, we adopt a Uniform Progress variant that first assigns spot instances locally and, if none are available, move to other regions in a round-robin manner.
We run OpenEvolve for 100 iterations with Gemini 2.5 Pro, using the same island setup as before. The evaluator then reports both overall and per-trace scores, similar to Section~\ref{sec:cant-be-late}.

\textbf{OpenEvolve results.}
The final policy achieves 26\% cost savings, on average, compare to the multi-region Uniform Progress baseline. It balances cost efficiency with deadline guarantees using a simple principle: when a job is not urgent (i.e., not at risk of missing its deadline), it explores additional regions to seek lower-cost spot capacity; if a job is urgent, it prioritizes immediate progress, selecting spot instances when available or falling back to on-demand. This adaptive logic enables opportunistic exploration under slack conditions while ensuring reliability when deadlines are at risk, effectively managing the trade-off between exploration and guaranteed progress in a multi-region environment. In addition, the policy leverages a dynamic view of regional capacity to opportunistically migrate when conditions are favorable.

\textbf{Evolution process.} The search process demonstrates iterative improvement of the deadline monitoring mechanisms from multi-region scheduling policies. 
Initial strategies implement basic progress tracking by comparing task completion against elapsed time.
The system discovers key insights through failure analysis.
At iteration 7, the system introduces region caching and urgency calculation. Iteration 5-12 attempts with aggressive cost reduction initially show promise, but ultimately fail when accumulated delays cannot be recovered within deadline constraints. These failures guide the search toward more balanced approaches.
The final evolved strategy at iteration 63 implements a two-stage urgency detection system. Rather than applying uniform resource allocation rules, it combines schedule-based progress monitoring with direct deadline pressure analysis. This design enables adaptive behavior: immediate allocation of on-demand instances when deadlines are at risk, while maintaining intelligent region exploration when deadline permits. The insight is the separation of deadline assessment from resource provisioning decisions, enabling adaptive region selection.





\subsection{Case Study \# 2: Optimizing Expert Placement in MoE Inference}
\label{sec:EPLB}

In this section, we discuss the problem of designing efficient algorithms to balance computational load during inference across multiple GPUs in a Mixture-of-Experts (MoE) architecture.
OpenEvolve discovers an algorithm implementation that runs 5.0$\times$ faster than the state-of-the-art frontier-lab reference implementation\shu{company reference implementation or frontier-lab reference implementation?} while achieving similar load balancing.
 
\textbf{Problem setup.} The basic Expert Parallelism Load Balancing (EPLB) algorithm runs in three stages: (i) distribute expert groups across nodes to balance the load, (ii) create replicas for hot (popular) experts, and (iii) assign these replicas to  GPUs to further minimize the imbalance. The problem is then: given a query workload, an MoE model and a set of GPUs, determine the number of replicas for each experts and then map these replicas on GPUs such that to minimize the imbalance.  

\textit{Objective and constraints.} 
Our optimization goal is twofold: to minimize load imbalance (i.e., the ratio of average to maximum tokens generated per GPU) and to reduce the running time of the algorithm used to re-balance experts when the load changes.

\textbf{Solution generator and selector.}
We run OpenEvolve with five islands, and we use a combination of 80\% Gemini 2.5 Flash and 20\% Gemini 2.5 Flash Lite. We cap the evolution at 300 iterations. The total evolution takes roughly five hours and costs less than \$10.


\textbf{Evaluator.} 
Our simulator models a distributed GPU inference engine for MoE models. The simulator is implemented in PyTorch and consists of 168 lines of code. Our evaluation trace models the load changes over the ShareGPT and GSM8K datasets~\cite{}.

\textit{Initial program and baselines.} The initial program as the open-source EPLB implementation from DeepSeek~\citeyearpar{deepseek-eplb}. This solution performs expert placement using a greedy bin-packing, i.e., it sorts experts by their load in descending order and assigns each to the least-loaded GPU that has capacity left. However, despite its simplicity, this solution is slow as it is written an Python and uses a for-loop to performs linear search for finding the best-fit GPU choice. On average, it takes about 540 ms to re-balance the experts and achieves an imbalance factor of 0.66.

As a baseline, we include a non-public reference implementation from a frontier lab to which we had access. It introduces a clever heuristic to replace the loop: instead of explicit bin packing, it reshapes and transposes tensors representing expert indices, using PyTorch’s fast tensor operations to effectively stagger expert assignments via a zigzag (or ``snake'') pattern between heavily and lightly loaded GPUs. The main idea is to alternate expert placement in a way that naturally interleaves high-load and low-load experts across GPU slots. This heuristic avoids explicit iteration and reduces the rebalancing algorithm runtime from 540 ms to 19.6 ms while achieving the same imbalance factor.\shu{Is it fine that we talk about implementation details from private code?}

\textit{Evaluator feedback.} The evaluator's ouput metrics are: (a) the imbalance factor and (b) the time it takes to rearrange the expert replicas when the load changes over our test datasets. Since OpenEvolve require us to provide a single metric, we compute this metric as the equally weighted average of the load imbalance factor and the rebalance algorithm runtime.




\begin{figure}[h!]
\caption{Side-by-side comparison of the initial greedy policy and final evolved heuristic. Key innovations in the evolved policy are highlighted.}
\label{fig:evolution_comparison}
\centering

\begin{subfigure}{0.45\textwidth}
    \begin{minted}[
        frame=lines,
        fontsize=\tiny,
        linenos,
        breaklines=true
    ]{python}
def InitialStrategy(...):
    ...
    for item in sorted(items, reverse=True):
        # Greedily choose least-loaded pack
        # Note that these are plain for-loops
        available_packs = filter_nonfull(packs)
        min_loaded_pack = min(available_packs)
        min_loaded_pack.add(item)
    ...
    \end{minted}
    \caption{Before: Initial Program}
    \label{fig:eplb-initial-code}
\end{subfigure}
\hfill 
%
\begin{subfigure}{0.45\textwidth}
    \begin{minted}[
        frame=lines,
        fontsize=\tiny,
        linenos,
        breaklines=true,
        highlightlines={13-17},
highlightcolor=blue!10
    ]{python}
def EvolvedStrategy(...):
    ...
    indices = torch.arange(num_items)
    block_id = indices // len(num_packs)
    idx_in_block = indices % len(num_packs)
    is_even_block = block_id % 2 == 0

    # Heuristics here: no loop at all!
    # The "snake" pattern:
    # For items in even blocks,
    # assign them to each pack while
    # for odd blocks, in a reversed order
    packs_for_sorted_items = torch.where(
        is_even_block,
        idx_in_block,
        num_packs - 1 - idx_in_block
    )

    # Now we have assigned a pack for each item
    update_packs(packs_for_sorted_items)
    ...
    \end{minted}
    \caption{After: Evolved Policy}
    \label{fig:eplb-evolved-code}
\end{subfigure}

\end{figure}



\textbf{OpenEvolve results.} 
The new algorithm generated by OpenEvolve independently discovers the staggered placement technique, learning to use tensor reshaping and reversal to evenly distribute expert load. 
Notably, our baseline implementation is not publicly available, making it highly unlikely that the Gemini models used in our experiments were exposed to this implementation during training. Impressively, OpenEvolve also introduces subtle enhancements, including improved ordering logic and more adaptive reshaping strategies. The resulting algorithm matches the imbalance factors of the baselines while reducing runtime to just 3.7 ms, yielding a 5.0$\times$ speedup over the internal reference implementation.

\textbf{Evolution process.}
OpenEvolve’s evolution trajectory can be characterized by two major steps in improving runtime: first, replacing Python for-loops with PyTorch tensor operations, and second, discovering the zigzag placement pattern (Figure~\ref{fig:eplb-evolved-code}). Interestingly, the initial introduction of the zigzag pattern did not yield immediate gains—the imbalance factor sometimes worsened, and rearrangement costs fluctuated. The breakthrough came later, when OpenEvolve learned to systematically reuse the zigzag partitioning heuristic across multiple stages of EPLB, rather than only in the initial group distribution. At this point, performance improved substantially, reducing rearrangement time by orders of magnitude.

The expert replication stage, by contrast, remained the most unstable throughout evolution. The system oscillated between strategies such as copying the least-used experts, overloading popular ones, or attempting proportional spreads. These experiments rarely improved the score, and ultimately the intuitive rule of replicating only overloaded experts prevailed. Consequently, many iterations were unproductive, with the main speed improvements coming from global reorganization logic that exploited PyTorch’s batched operations and the zigzag layout.

In summary, OpenEvolve independently rediscovered and fully exploited a tensorized zigzag partitioning scheme, yielding an evolved EPLB algorithm that achieves a 5.0$\times$ speedup without worsening the imbalance factor.

\subsection{Case Study \#3: Optimizing LLM Inference on SQL Queries [MlSys `25]} 
This research problem~\cite{liu2024optimizing} arises in relational analytics, where SQL queries invoke LLMs over entire tables, with each row triggering a separate LLM inference operation. At scale, this is prohibitively expensive. The state-of-the-art solution mitigates cost by reordering rows and fields to maximize prefix KV cache reuse. 
Using OpenEvolve, we evolve such a reordering policy, achieving similar hit rates while delivering a 3$\times$ runtime speedup.

\begin{figure}[h!]
\centering
\caption{Side-by-side comparison of the greedy recursive grouping (QuickGreedy) and the evolved prefix-aware reordering policy. Key innovations in the evolved algorithm are highlighted.}
\label{fig:reorder_comparison}

\begin{subfigure}[t]{0.48\textwidth}
\begin{minted}[
frame=lines,
fontsize=\tiny,
linenos,
breaklines=true
]{python}
def QuickGreedy(df):
    # 1. Compute value counts for all cells
    counts = Counter(df.stack())
    val_len = {v: len(str(v))**2 for v in counts}

    # 2. Pick value maximizing len^2 * (count-1)
    v_star = argmax_v [ val_len[v] * (counts[v]-1) ]
    if v_star is None: return fixed_reorder(df)

    # 3. Split rows with/without v_star
    G = rows with v_star
    R = rows without v_star

    # 4. Reorder columns in G (v_star front, deps after)
    for row in G:
        cols_with_val = [c for c in df.columns if row[c]==v_star]
        reordered = cols_with_val + (others)
        row = row[reordered]

    # 5. Recurse on G remainder and R
    G = QuickGreedy(G remainder)
    R = QuickGreedy(R)
    return concat(G,R)
\end{minted}
\caption{QuickGreedy baseline.}
\label{fig:quick_greedy_code}
\end{subfigure}\hfill%
\begin{subfigure}[t]{0.48\textwidth}
\begin{minted}[
frame=lines,
fontsize=\tiny,
linenos,
breaklines=true,
highlightlines={6-8,12-15,23-32},
highlightcolor=blue!10
]{python}
def EvolvedPolicy(df):
    # 1. Cache value lengths once
    counts = Counter(df.stack())
    val_len = {v: len(str(v))**2 for v in counts}

    # 2. Larger base cutoff → fewer recursions
    BASE = 4000

    # 3. Find max group value using cached counts
    v_star = argmax_v [ val_len[v] * (counts[v]-1) ]
    if v_star is None: return fixed_reorder(df)

    # 4. Recurse only if |df| > BASE
    if len(df) > BASE:
        top, bottom = split(df)
        return concat(EvolvedPolicy(top),
                      EvolvedPolicy(bottom))

    # 5. Lightweight per-row heuristic
    # Columns equal to previous row first,
    # sorted by squared length, then others
    prev = None
    for row in df:
        if prev is None: order = df.columns
        else:
            matches = [c for c in df.columns if row[c]==prev[c]]
            matches.sort(key=lambda c: val_len[row[c]],
                         reverse=True)
            order = matches + [c for c in df.columns if c not in matches]
        prev = row
        row = row[order]

    return df
\end{minted}
\caption{Evolved prefix-aware policy.}
\label{fig:evolved_policy_code}
\end{subfigure}

\end{figure}

\textbf{Problem setup.} To minimize inference time and cost, we aim to maximize the prefix cache hit rate (PHR) by reordering both rows and fields in the table before performing inference. This problem is combinatorial: for a table with $n$ rows and $m$ fields, there are $n! \times (m!^n)$ possible orderings, making a naive brute-force search infeasible. Thus, the goal is to design a reordering algorithm that achieves high PHR while keeping its runtime small relative to the overall inference time.

\textit{Objective and constraints.} Our objective is to maximize the prefix hit rate (PHR) while keeping the runtime of the reordering algorithm low. Since PHR measures the fraction of token prefixes shared across consecutive rows, and serves as a proxy for inference cost and latency.


\textbf{Solution generator and selector.} We configure OpenEvolve with three islands, and we use an LLM ensemble of 80\% OpenAI o3 and 20\% Gemini 2.5 Pro, and we run it for 100 iterations. The entire evolution takes about one hour and costs less than \$7.

\textbf{Evaluator.} We leverage the publicly available simulator from the paper that measures prefix cache hit rate (PHR) given dataset table. 
The simulator is written in Python (200 LOC) and evaluates a benchmark of representative LLM queries across five recommendation datasets (e.g., movies~\cite{rotten-tomatoes-movies-dataset}, beer~\cite{ratebeer}, BIRD~\cite{bird}, PDMX~\cite{pdmx}, products~\cite{amazon-product-review-dataset}).

\textit{Initial program and baselines.} 
As initial program, we use the greedy recursive group algorithm (GGR)~\cite{liu2024optimizing}, a heuristic that recursively groups rows by common field values and reorders fields using schema statistics with early stopping. This approach approximates the optimal reordering algorithm while running more efficiently. For comparison, we also include a simple baseline: the table in its original ordering, i.e., the default row/field order of the input table. This program is implemented in Pandas~\cite{mckinney2010data}, an open-source Python library for data manipulation and analysis.

\textit{Evaluator feedback.} 
The evaluator reports four key metrics. First, it reports a combine score $ = 0.5 \times \text{PHR} + 0.5 \times \frac{1}{1 + \text{runtime}}$, defined as the equally weighted average of PHR and algorithm runtime across datasets, where higher PHR and lower runtime yield a higher score. To preserve query semantics, the reordering algorithm must not alter which rows or fields are included, only their order. Second, the evaluator records a binary flag indicating whether the candidate program executes end-to-end. Third, it reports the detailed prefix hit rate for each dataset. Finally, it measures the total runtime of the policy. The combined score serves as the optimization objective during evolution.

\textbf{OpenEvolve results.} The new program generated by OpenEvolve achieves a similar average PHR compared to the GGR algorithm, while reducing runtime by 3$\times$, thereby yielding a higher combined score. In contrast, the greedy baseline spends most of its time recomputing value counts and recursively traversing the table, which results in high overhead. The evolved solution introduces several optimizations. First, instead of recomputing full value counts at each recursion, it maintains a lazily updated global frequency map, eliminating redundant data traversals. Second, it replaces slow Pandas lookups with a direct attribute-mapping method, reducing the core loop from costly Pandas calls to straightforward $O(N_\text{rows} \times N_\text{cols})$ Python operations. Finally, it applies a local heuristic for per-row ordering: instead of globally sorting the entire table, it reorders fields by maximizing continuity with the preceding row while weighting by squared value length. These features are highlighted and described in Figure~\ref{fig:reorder_comparison}.

\textbf{Evolution process.} 
The search begins with the published GGR algorithm, which achieves a good Prefix Hit Rate (PHR) but suffers from repeated counter operations and deep recursion. Early in the search, by iteration 32, OpenEvolve discovers a faster heuristic that orders columns by (frequency $\times$ squared length) instead of using recursive splitting. This change eliminates redundant value counting and improves runtime, though at the cost of some grouping accuracy.

Later, by iteration 72, the heuristic is refined. It now uses normalized weights (frequency ratio $\times$ squared length) and limits multi-key sorting to only the most informative columns, which improves hit rates while maintaining efficiency.

By iteration 97, the final program strikes an optimal balance between speed and accuracy. It raises the recursion base threshold, reuses cached counts and string lengths, reintroduces selective recursion for rows, and incorporates a NumPy-based reordering to replace costly pandas lookups.

Overall, the evolution progresses from early runtime improvements to mid-stage refinements that recover accuracy, culminating in a final design that integrates both for the best overall score.

\subsection{Case Study \#4: Optimizing Transaction Scheduling [VLDB `24]}
\label{sec:transaction-scheduling}

This research problem~\cite{cheng2024towards} aims to find efficient schedules to reduce conflicts for transactional workloads. Transaction processing systems can significantly improve throughput by determining and executing transactions in an order that minimizes overall execution time. We apply OpenEvolve to this problem and find that the framework is unable to find a more successful algorithm than the existing state-of-the-art policy when constrained to the online setting as in the original problem. However, OpenEvolve is able to find an algorithm that provides better schedules without online constraints, demonstrating that it can be useful for rapidly exploring different variations of problems.

\textbf{Problem setup.} Conflicts on shared data cause performance bottlenecks in many transactional workloads~\cite{taobench}. One approach to minimize conflicts is to carefully schedule the transactions. The problem we aim to solve is: given a set of transactions, find a schedule that minimizes the conflicts and improves the throughput. 

\textbf{Solution generator and selector.} 
We configure OpenEvolve to run with three islands and use a two-model ensemble of 80\% Gemini 2.5 Pro and 20\% OpenAI o3 for both problem settings. We run for 100 iterations, which takes less than two hours and costs less than \$20. 

\textit{Objective and constraints.} Maximizing throughput in this setting is equivalent to minimizing the schedule \textit{makespan}, i.e., the total time to execute all transactions. We consider both the online and the offline settings for this problem. In the online setting, we assume that the transaction order is fixed once the schedule is determined (i.e., committed transactions cannot be rollbacked). We constrain the scheduling algorithm to O($n$) runtime (where $n$ is the number of transactions to be scheduled) and assume the read/write operations are not known apriori (only hot keys can be predicted). We also consider the offline scheduling problem, which is relevant to deterministic databases~\cite{} that schedule batches of transactions, a setting with no previously known results.

\textit{Initial program and baselines.} We use the state-of-the-art algorithm, Shortest Makespan First (SMF), which greedily chooses transactions to schedule, as our initial program. 
We also compare against a number of transactional scheduling algorithms as well as a simple baseline that picks the transactions at random.

\textit{Evaluator.} We use the Python simulator from the SMF paper~\cite{cheng2024towards}, which assumes that each operation takes one unit of time. The simulator calculates the makespan of a given transaction schedule and also provides statistical bounds on the makespan of the schedule for a given workload. We measure total makespan over five traces from the OLTP benchmarks used in the original paper (Epinions, SmallBank, TPC-C, TAOBench, YCSB) with 500 transactions each. We use the random-scheme baseline as the initial program. 


\textbf{OpenEvolve results.} In the online setting, the best discovered policy is SMF.
We note that OpenEvolve is able to rediscover this algorithm from a random scheduling baseline.
It is likely that this is a case of contamination, i.e., the model was trained upon the SMF paper. While this result is not as interesting, it shows these frameworks can reproduce state-of-the-art solutions. Furthermore, it is not unexpected that OpenEvolve is unable to find a better solution in this case: transaction scheduling is a complex optimization problem, where we need to consider groups of operations, dependencies across operations, and correctness constraints (e.g., serializability). Prior work found that more involved heuristics, such as hill climbing, simulated annealing, etc., did not perform better due to the complexity of the problem constraints~\cite{cheng2024towards}. Instead, leveraging the cost of conflicts, as SMF does, provides a generalizable solution that works across different workloads.

At a high level, the intuition behind SMF is to minimize the cost of conflicts as the schedule is constructed by placing transactions with high conflict costs far apart. The incremental \textit{makespan} increase when a given transaction is added to the schedule accounts for the cost of all potential conflicts that an unscheduled transaction has with the current ordering. Concretely, SMF starts the schedule with a random transaction and at each iteration (Figure~\ref{fig:smf}, lines 2-3), finds the transaction that increases makespan the least among $k$ randomly sampled unscheduled requests (lines 5-25) and append this transaction to the schedule (lines 28-35). For tie-breaking, a transaction is randomly chosen. It has a linear runtime of $O(n \times k)$ run time, where $n$ is the number of transactions to be scheduled and $k$ is a constant representing the sample size. 



In the offline setting, OpenEvolve discovers a novel algorithm than reduces makespan by 34\% compared to SMF. This result reduces concerns about contamination since it is not a previously known solution. The final algorithm involves three parts. First, it computes simple features per transaction (number of writes, length) and builds a strong initial sequence by sorting transactions lexicographically by (fewest writes, then shortest), which tends to reduce conflicts (Figure~\ref{fig:txn_offline}, lines 6-13). Second, it then runs the full greedy algorithm in which each transaction is tried in every position (lines 16--33) It  then tries to optimize for any local minima due by performing a pair-swap hill climb for a fixed number of iterations. Finally, it tries a few random schedules as a safety net (lines 49--55). This algorithm extends the greedy intuition of SMF and has $O(n^2)$ runtime. This result indicates that scheduling based on cost of conflicts is the right approach to reduce makespan. Furthermore, OpenEvolve is able to quickly develop a variation of the algorithm for altered problem constraints (e.g., no runtime or reordering constraints), showing it can assist researchers in developing solutions for different problem settings (that may otherwise require manual algorithm re-design).


\begin{figure}[h!]
\centering
\caption{Side-by-side comparison of the evolved constant-time greedy policy (SMF) and the offline evolved policy. Key differences in the offline policy are highlighted.}
\label{fig:schedule_comparison_numbered}

\begin{subfigure}[t]{0.47\textwidth}
\begin{minted}[
frame=lines,
fontsize=\tiny,
linenos,
breaklines=true
]{python}
# --- 1. Candidate sampling ---------- #
SAMPLE_SIZE = 8  
rng = np.random.default_rng()

while remaining:
    # --- 2. Draw candidates --------------- #
    # Pick a constant-sized random subset
    if len(remaining) <= SAMPLE_SIZE:
        candidates = remaining
    else:
        candidates = rng.choice(remaining, size=SAMPLE_SIZE, replace=False)

    # --- 3. Evaluate incremental cost --- #
    best_cand = None
    best_extra = math.inf
    best_start_end = (0, 0)

    for cand in candidates:
        extra_cost, t_start, t_end = compute_incremental_cost(
            self.txns[cand], key_map, txn_id, total_cost
        )
        if extra_cost < best_extra:
            best_extra = extra_cost
            best_cand = cand
            best_start_end = (t_start, t_end)

    # --- 4. Commit update global state ----- #
    schedule.append(best_cand)
    remaining.remove(best_cand)

    apply_txn_to_keymap(
        self.txns[best_cand], key_map, txn_id, best_start_end[0]
    )
    total_cost += best_extra
    txn_id += 1

return total_cost, schedule
\end{minted}
\caption{SMF / greedy policy.}
\label{fig:smf}
\end{subfigure}\hfill%
\begin{subfigure}[t]{0.47\textwidth}
\begin{minted}[
frame=lines,
fontsize=\tiny,
linenos,
breaklines=true,
highlightlines={6-8,11-13,19,36-46,49-55},
highlightcolor=blue!10
]{python}
def get_best_schedule(self, num_seqs: int = 10):
n = len(self.txns)
idxs = list(range(n))

# --- 0. Pre-compute transaction statistics --- #
write_cnt = [
    sum(1 for op in self.txns[t] if op[0] == "w") for t in idxs
]

# - 1. Sorted seed sequence (by #writes, length) - #
base_seq = sorted(idxs, key=lambda t: (write_cnt[t], len(self.txns[t])))
best_seq = base_seq
best_cost = self.get_opt_seq_cost(best_seq)

# --- 2. Greedy insertion ------- #
improved = True
while improved:
    improved = False
    for i in range(n):
        txn = best_seq.pop(i)
        insert_best_pos = 0
        insert_best_cost = float("inf")
        for pos in range(n):
            candidate = best_seq[:pos] + [txn] + best_seq[pos:]
            cost = self.get_opt_seq_cost(candidate)
            if cost < insert_best_cost:
                insert_best_cost = cost
                insert_best_pos = pos
        best_seq.insert(insert_best_pos, txn)
        if insert_best_cost < best_cost:
            best_cost = insert_best_cost
            improved = True
            break

# --- 3. Pair-swap hill climb -------- #
max_iters = max(100, num_seqs * 50)
for _ in range(max_iters):
    i, j = random.sample(range(n), 2)
    if i == j:
        continue
    new_seq = best_seq.copy()
    new_seq[i], new_seq[j] = new_seq[j], new_seq[i]
    new_cost = self.get_opt_seq_cost(new_seq)
    if new_cost < best_cost:
        best_cost = new_cost
        best_seq = new_seq

# --- 4. Random restarts (keep best) ----------- #
for _ in range(num_seqs):
    seq = idxs[:]
    random.shuffle(seq)
    cost = self.get_opt_seq_cost(seq)
    if cost < best_cost:
        best_cost = cost
        best_seq = seq

return best_cost, best_seq
\end{minted}
\caption{Offline evolved policy.}
\label{fig:txn_offline}
\end{subfigure}

\end{figure}

\textbf{Evolution Process} 
For the online setting, OpenEvolve shifts from random schedules and length-only heuristics toward increasingly conflict-aware solutions. As the framework recognized that contention leads to higher makespan, it explores different heuristics, such as write-count bucketing and greedily constructing schedules to minimize the contentions. Some common pitfalls in the search process were over-reliance on transaction length (length correlates poorly with contention) and full greedy selection that violated the O(n) constraint. 
The search often showed early over-confidence in single heuristics (length-first, then write-first), converging quickly but missing gains from other techniques. The best program generalized the evolution’s heuristic insights (account for the cost of conflicts) into the state-of-the-art policy.

For the offline setting, the evolution also shifts from random sampling to simple heuristics (e.g., shortest-first, fewest-writes). The search eventually centered in two directions: (i) greedily appending transactions to minimize makespan at each iteration and (ii) metaheuristics that broadened exploration using simulated annealing (SA) and neighborhood swaps. The search often got stuck when swap-only hill climbs plateaued quickly in local minima when starting from random schedules. The framework later discovered that pairing this technique with greedy construction led to better schedules. In particular, the reasoning models demonstrated a better understanding of how to escape local minima by using problem-relevant signals (e.g., key frequency) and considering larger neighborhoods (e.g., insert into every possible position). The best solution combined the successful techniques that emerged during evolution: a cheap yet informative heuristic start (sorting transactions based on writes/length), a powerful greedy construction method, and a lightweight swap hill climb plus a few random restarts for diversification.
\section{Early Best Practices}
\label{sec:best-practices}

Although existing \SYS/ frameworks have demonstrated potential in our case studies, these systems are still in their infancy. They exhibit several significant limitations, such as the size and complexity of the code they can analyze, and are prone to failures. Table~\ref{tab:f} provides a taxonomy of the most common failures we have encountered, which fall into three main categories:

\begin{itemize}

\item \emph{Runtime Errors.} The generated code fails to run due to compilation errors or exceeding the experiment's resource budget.

\item \emph{Search Failures.} The evolutionary process stalls, such as getting stuck in a local optimum or alternating between poor solutions.

\item \emph{Algorithm Failures.} The final solution runs, but is flawed or does not improve the baseline. This includes solutions that ignore problem constraints, exploit loopholes in the evaluator, or rely on shallow tweaks.

\end{itemize}

We discuss these failures in more detail in Appendix~\ref{sec:failure-taxonomy}. 
Based on our experience, next, we present the best-practices to avoid these failures and alleviate other limitations. We group these gudelines by the main components of \SYS/, as depicted in Figure~\ref{fig:evolve-sys}.

\begin{table}[h!]
\centering
\renewcommand{\arraystretch}{1.35} 
\setlength{\tabcolsep}{6pt} 
\caption{Common failure patterns in \SYS/ pipelines (distribution estimated from 420 LLM-judged traces).}
\label{tab:f}
\resizebox{\linewidth}{!}{%
\begin{tabular}{p{2.8cm} l p{8.8cm}}
\toprule
\textbf{Category} & \textbf{Failure Type} & \textbf{Description} \\
\midrule
\multirow{2}{*}{\makecell[l]{Runtime\\Errors}} 
& \textit{Syntax \& Interface Errors} & Candidate solution fails to compile or integrate with evaluator. \\
& \textit{Budget Exhaustion} & Candidate exceeds resource limits (e.g., context window, API quotas, timeouts). \\
\addlinespace
\midrule
\multirow{3}{*}{\makecell[l]{Search\\Failures}} 
& \textit{Premature Convergence} & Search settles on a local optimal solution too early. \\
& \textit{Stuck-in-the-Loop} & Search repeats similar solutions without meaningful progress. \\
& \textit{Mutation Drift} & Search produces contradicting or random edits to the solution. \\
\addlinespace
\midrule
\multirow{4}{*}{\makecell[l]{Algorithm\\Failures}} 
& \textit{Misaligned Objectives} & Solutions ignore key constraints (e.g., latency SLOs). \\
& \textit{Sub-Optimal Optimizations} & Shallow changes (e.g., API calls) instead of substantive algorithmic improvement. \\
& \textit{Overfitting} & Hard-coded / narrow solutions underperform on unseen traces. \\
& \textit{Reward Hacking} & Solution exploits loopholes in the evaluator rather than solving intended problem. \\
\bottomrule
\end{tabular}
}
\end{table}

\subsection{Prompt Generator}
A clear and well-scoped problem formulation is the foundation of effective algorithm evolution. 

\textbf{Provide structured specifications.} 
Many execution and algorithm failures trace back to missing context, such as critical details about the problem or absent code API documentation. 
A well-designed prompt should be as specific and structured as possible, clearly defining three key areas:

\begin{itemize}[noitemsep,topsep=0pt,parsep=0pt,partopsep=0pt]
\item \emph{The problem:} what is the core task to solve.
\item \emph{The evaluation criteria:} how a solution will be evaluated, including optimization goals and correctness constraints.
\item \emph{The context:} any necessary information, such as required APIs. 
\end{itemize}

We recommend drafting prompts with external LLMs (e.g., ChatGPT, Gemini, etc.) to craft a structure before launching evolution. 

\textbf{Provide a suitable base program.}  
The choice of base program strongly shapes the trajectory of algorithm evolution. Buggy or weak baselines waste iterations on trivial fixes (\textit{budget exhaustion}), while a strong, clean baseline can accelerate progress toward meaningful improvements. 
For example, in the LLM-SQL case study, the published baseline already achieved near state-of-the-art prefix hit rate; the main bottleneck was runtime, so $\sim$100 iterations sufficed to evolve a solution that was $3\times$ faster without loss in PHR.  

Conversely, overly strong baselines that encode near-SOTA solutions or rely on high-level APIs can limit the search to shallow \textit{micro-optimizations}. In the Can't-be-Late problem (Section~\ref{sec:cant-be-late}), evolution from a simple greedy baseline produced better results than starting from the stronger Uniform Progress policy, which restricted exploration. We recommend seeding evolution with clean, minimal, high-quality baselines, e.g., using coding assistants such as Claude Code.

\textbf{Provide suitable solution hints.} 
While a detailed problem specification is always beneficial, the value of providing solution hints -- specific suggestions for how the system should approach the problem -- is more nuanced.
Too much guidance can risk \textit{premature convergence} and prevent the discovery of novel solutions, while too little can make the search inefficient (i.e., \textit{stuck-in-the-loop}).
For example, in the EPLB problem, hints could have prevented wasted iterations on “extreme” replication strategies. However, in the transaction scheduling use case, hints about batching biased the search toward sub-optimal designs, whereas leaving it unconstrained led to a 30\% faster greedy policy in OpenEvolve. 

We find that providing intermediate human feedback as hints is especially effective when the search gets stuck in the loop. In summary, we recommend trying several prompts with different levels of hint specificity and inject relevant hints as how the evolution progresses.



\textbf{Choose a suitable level of abstraction.} We recommend exposing only the level of abstraction that matches your goal.
Allowing full access to high-level external library APIs can sometimes lead to \textit{sub-optimal optimizations}: e.g., trivial speedups from replacing custom operators with PyTorch primitives, while blocking deeper innovation.
To encourage algorithmic advances, restrict API access to help the system explore new strategies rather than rely on pre-built solutions. On the other hand, when the goal is execution efficiency, providing optimized library access is appropriate.
In practice, tuning this boundary between enabling useful shortcuts and enforcing genuine problem-solving is critical to avoid micro-optimizations.

\subsection{Solution Generator}
\textbf{Use model ensembles.} 
Search failures often arise when the solution generator either over-explores or over-exploits.
We recommend using an ensemble of models to balance exploration and exploitation.
On one hand, reasoning models, such as o3, encourage exploration by generating novel solutions. 
In our ablation study for the `Can't Be Late'' problem (Section~\ref{sec:spot-instances}), o3 provided more creative ideas that led to higher-performing solutions. 
However, these models are typically slower and more expensive.
On the other hand, non-reasoning models are effective at refining existing solutions more efficiently. This enables OpenEvolve to iterate faster and perform many more iterations at the same budget. 

Therefore, we currently recommend using two different models for efficient exploration. 
We find out that using more than two models often introduces conflicting ideas, leading to instability and \textit{mutation drift} to the candidate solutions. 



\subsection{Evaluator}

\textbf{Prevent overfitting.} Evaluating against narrow workloads lead to algorithm failures like \textit{overfitting}, where the solutions either hard-code behaviors or overfit to specific traces. 
For example, our ablation study in the Can't-be-Late use case (Appendix~\ref{appendix-1}) shows that limited spot scheduling evaluation to a single availability pattern led to worse performance on workloads that are not part of the evaluation set. 
Broader and more diverse test sets that cover edge cases help mitigate this issue.


\textbf{Prevent reward hacking.} 
Reward hacking occurs when solutions exploit evaluator loopholes rather than solving the intended problem. For example, when we evolve code for detecting failures in a multi-agent system, some candidates bypassed one stage of the evaluation pipeline, achieving high scores without performing the full task. 
To avoid this, evaluators should combine multiple evaluation signals (e.g., correctness, efficiency, robustness) and add adversarial tests.
This makes it harder for \SYS/ frameworks to "game" the evaluation and also reduces failures of \textit{misaligned objectives}, ensuring candidates satisfy all constraints.


\subsection{Solution Selector}
Search outcomes also depend heavily on how candidates are retained between iterations, which is determined by the solution selector.

\textbf{Balance exploration-exploitation.} Greedy selectors converge quickly but risk \textit{premature convergence}, while overly random ones waste resources and cause \textit{stuck-in-the-loop} failures. 
Effective selectors maintain diversity while still advancing higher-quality candidates. 
For example, AlphaEvolve blends MAP-Elites with island-based evolution to strike this balance. 
In practice, the exploration–exploitation ratio is a tunable parameter (e.g., in OpenEvolve), and careful adjustment is key to avoiding search failures. 

\section{Limitations and Open Challenges}

In this section, we discuss the limitations of \SYS/—specifically, the types of problems for which the \SYS/ approach is well suited and those for which it is not—and then outline several open challenges aimed at improving \SYS/ so that it can successfully address an increasingly broader range of problems.

\subsection{Which Problems Are Best Suited for \SYS/, and Which Are Not?}

\label{sec:limitations}

Based on our early experience, \SYS/ works best for problems that require localized changes, are fast to evaluate, and easily verifiable.
It is less effective for problems that require changes across many systems, rely on weak verifiers, or involve costly evaluations. 
Understanding these boundaries will help researchers apply \SYS/ where it is most likely to succeed and adapt systems to better align with its strengths. 
In summary, \SYS/ is well suited for systems performance problems that exhibit the following properties:
\begin{itemize}

\item \emph{Isolated changes:} Since existing LLMs are more reliable when modifying small amounts of code, \SYS/ is a better fit for problems that require improving policies or algorithms that are isolated from the rest of the system. Examples include schedulers, cache managers, load balancers, and resource allocators. In contrast, existing \SYS/ systems do not work well for optimizing policies or protocols that are distributed across large systems, such as consensus protocols that involve a complex interplay between components (e.g. state management, network communication and failure detection logic~\cite{paxos,raft}).
\mert{Is there an example where someone tried to optimize such a ``complex-to-evolve" system with OpenEvolve and failed? If so, it could be good to elaborate on precisely how open-evolve gets stuck might be nice.}
\shu{sort of agree with Mert: is there some reason why it does not work? is it coding capability limitations? is it context limitation? } \ion{OpenEvolve does changes in a single file; multiple files are inherently harder; require a larger content and that's less accurate as most of the LLMs right now use sparse attention; their performance might degrade as the context is increasing.} \melissa{I'd treat single vs. multi file an implementation limit of OpenEvolve, but very large real-world systems (Terabytes of production system code) a real limitation for current LM.}

\item \emph{Reliable evaluations:} Given two solutions, it should be easy to determine which one performs better and whether the two are semantically equivalent, i.e., they produce the same outputs. One example is improving a load balancer: it is straightforward to determine which solution is better by measuring the imbalance factor, and any two solutions are trivially semantically equivalent if every query is routed to a replica of the same service. However, in other cases, proving semantic equivalence can be difficult or even infeasible. For instance, arbitrarily modifying a query plan and verifying that it is equivalent to the original plan is, in general, impossible~\cite{AbiteboulHullVianu1995}, though certain transformations provably preserve semantics, such as reordering joins or reordering columns or rows in a table.\shu{does reliable evaluations == how easy it is to verify correctness of the solution? Find it a bit hard to read here}

\item \emph{Efficient evaluations:} The evaluation must be both fast and cost-effective. Using \SYS/ to discover new solutions may require hundreds or even thousands of iterations. If each evaluation takes many hours to complete and costs hundreds of dollars, the overall process can take months or even years and amount to hundreds of thousands of dollars. Some examples are GPU-intensive workloads, such as weight compression, which can take more than 12 hours per evolution cycle (see Section~\ref{sec:case_studies}), or large-scale distributed training across thousands of GPUs. 
\end{itemize}

Finally, we do not expect \SYS/ to be as effective for problems that can be cast as optimization problems and solved directly using existing solvers, such as integer linear programming (ILP) solvers.

\subsection{Open Challenges}
\label{sec:open_challenges}
As \SYS/ is emerging as a promising approach to accelerate systems research, two natural questions arise: what should we build to better support \SYS/ and how should we improve \SYS/ itself?

\subsubsection{Supporting \SYS/: Build Better Evaluators}
\label{sec:better-evaluators}

A recurring lesson from our case studies is that \SYS/ is only as good as the evaluators guiding it. Successful discovery requires \emph{evaluators} that provide accurate, fast, and detailed feedback to the search process. We distill below the key properties that should guide the design of evaluators for \SYS/:

\begin{itemize}

\item \emph{Fidelity:} Evaluators must capture the salient behaviors of the system that are relevant to the problem being solved (e.g., flow-level or packet-level fidelity in networking). The desired level of fidelity depends not only on the system under study, but also on the type of solution that \SYS/ might explore.

\item \emph{Generality:} Evaluators must support diverse workloads and traces to provide reliable feedback signals and prevent overfitting to a single configuration or dataset.

\item \emph{Speed and reliability:} As discussed above, reliable and efficient evaluators are key to the successful application of \SYS/. Building testbeds that support rapid forking and rollback (e.g., via lightweight VMs, container snapshots, or database cloning) can drastically shorten feedback cycles and improve \SYS/ scalability~\cite{liu2025supporting}.

\end{itemize}

Next, we suggest two approaches to improve the efficiency of the evaluation framework.

\textbf{Problem-specific simulators.} Simulators offer fast, low-cost evaluation but are hard to design, as they must balance fidelity and simplicity for effective LLM reasoning.
A practical solution is to build domain-specific simulators that capture only behaviors relevant to the task: for instance, modeling CPU scheduling without full operating system details.
Tools like OpenEvolve can iteratively refine these simulators until their metrics (e.g., latency, throughput) align with real systems. Though costly to build, such simulators can be reused across related problems, amortizing development effort.

\textbf{Cascading evaluators.} 
Fast and accurate evaluation can be achieved by cascading evaluators: from fast, coarse-grained cost models to slower but high-fidelity ones. 
For example, one might begin with a simple cost model (e.g., for database queries~\cite{SiddiquiJindalQiaoPatelLe2020}), then progress to simulators of increasing granularity, followed by emulators, and finally tests on the real system. In networking, for instance, session-level simulators offer coarser evaluation compared to packet-level simulators. This mirrors standard research practice: prototype quickly, then validate precisely. 

\subsubsection{Improving \SYS/}

To improve the reach of \SYS/, we need advances across the following key components. 

\textbf{Prompt generator.}
Existing \SYS/ frameworks often rely on relatively simple prompts that provide only a problem description.
Just as human researchers ground their ideas in prior work, \SYS/ frameworks should incorporate retrieval techniques~\cite{packer2023memgpt} to draw from a broader body of knowledge (e.g., academic literature, documentation, and related examples) to better guide their search.
As discussed in Section~\ref{sec:best-practices}, providing richer solution hints in a prompt improves search efficiency, while fewer hints may encourage broader exploration. 
Prompt evolution~\cite{alphaevolve, agrawal2025gepa}, where the LLM refines its own context, can help manage this trade-off.

\textbf{Solution generator.}
\label{sec:future-work-sol-generator}
A strong solution generator should act like a autonomous developer, being able to navigate, reason over, and modify a system codebase rather than isolated code snippets. 
For example, modern AI workloads often rely on workload analysis and cross-layer optimizations and to reduce hardware costs~\cite{chung2024toward}. Similarly, distributed systems that optimize communication protocols may require coordinated changes to both the sender and the receiver logic~\cite{raft,opaxos}. 
Achieving this demands more agentic coding abilities from the solution generator: understanding dependencies, invoking analysis tools, and reasoning coherently across non-contiguous code modules.

Beyond a single model, \SYS/ frameworks should also support ensembles of specialized agents that collaborate to produce high-quality solutions.
Future work should focus on developing tools that automatically optimize the composition based on the given problem, much like forming a research team with complementary skills.

\textbf{Evaluator.}
Some of the existing \SYS/ frameworks like OpenEvolve require researchers to formalize intuitive trade-offs into numerical weights, which can be difficult in practice. For example, in our EPLB case study, we struggled to decide exactly how to weigh the importance of load balance against the re-balancing algorithm runtime.  Future systems could instead \textit{learn} user preferences automatically. Some possible approaches are: 
\begin{itemize}

\item \emph{Preference Learning.} \SYS/ could present researchers with pairs of solutions (e.g., “Solution A is faster but less fair; Solution B is fairer but slower”), and the researcher selects the one they prefer. By observing these choices, \SYS/ can infer the underlying objective function to optimize.

\item \emph{Inverse Reinforcement Learning (IRL).} If a researcher can provide a few examples of “good” solutions (e.g., those with particular levels of fairness and performance), \SYS/ can work backward to infer the reward function most likely to generate such solutions. 




\end{itemize}

\textbf{Solution selector.}
The process by which \SYS/ frameworks discover new solutions remains a \emph{black box}. 
Current evolutionary searches are monolithic, producing only a single final program, often mixing a good idea with poor implementation and thus penalizing promising concepts.
Prior work~\cite{tang2025ai} suggests separating ideation from code generation to address this issue.

The evolutionary search is also inefficient, frequently looping over failed heuristics or repeated errors (Appendix~\ref{sec:failure-taxonomy}). More flexible frameworks are needed to allow finer-grained feedback, letting humans or LLMs lock in working code, boost diversity to escape local optima, or roll back to prior versions when evolution stalls.

\textbf{Overall framework.}
\label{sec:future-work-framework}
Finally, we discuss open challenges related to the overall \SYS/ approach. 

\textit{Hyperparameter tuning.} Balancing exploration and exploitation in LLMs remains difficult and often relies on trial and error. Future work should focus on automating this tuning process. A meta-learning layer, for instance, could allow the system to learn and adjust as evolution progresses, making the entire framework more reliable and accessible.

\textit{Human-\SYS/ interaction.} The optimal balance between synchronous (interactive assistants like Cursor) and asynchronous (autonomous frameworks like OpenEvolve) user interface remains an open question. A key challenge remains for \SYS/ is to define when human guidance adds values versus when \SYS/ should act autonomously. 


\section{How Can \SYS/ Impact the Research Process?}
\label{sec:esearch-process-impact}

Despite its limitations (see Section~\ref{sec:limitations}), \SYS/ can already help researchers in two key ways:

\begin{itemize} 

\item \emph{Accelerating discovery:} 
\SYS/ tools automate tedious tasks such as implementation and, to some extent, debugging. This frees researchers to focus on problem selection, system architecture and design. Even imperfect solutions offer value by revealing new directions. In our telemetry repair case study, AI-generated insights shaped a better human-designed algorithm.

\item \emph{Achieve better-than-human results:} \SYS/ tools can explore the solution space more thoroughly than humans. Where a human might stop after a breakthrough, AI can continue testing variations for incremental gains. These small improvements compound. In our EPLB case study, OpenEvolve produced an algorithm that surpassed the state-of-the-art by exploiting missed optimizations.

\end{itemize}

As such, \SYS/ has the potential to reshape systems research as we know it. 

One way to think about \SYS/ is as providing researchers with a virtually unbounded number of ``assistants'' that follow directions reliably. Much like junior researchers, \SYS/ frameworks are most effective when given clear problem specifications and well-defined goals.

As a result, \SYS/ adds another layer to an already established research hierarchy. In academia, for example, a faculty member typically advises Ph.D. students or postdocs, who in turn mentor undergraduate students. Each of these roles can now leverage \SYS/ tools to accelerate their work and broaden the scope of what they can accomplish.

In this context, two natural questions arise:

\textbf {How would the role of a researcher change?} As \SYS/ tools begin to solve an increasing number of problems autonomously, researchers will have more time to focus on higher-leverage activities—selecting which problems to pursue and formulating them precisely. 
If successful, \SYS/ has the potential to elevate everyone in the research hierarchy—faculty and students alike—and make them far more productive.

\textbf{Will \SYS/ lead to fewer researchers?} We believe the opposite is true. If anything, \SYS/ will expand the research community by enabling individuals who may not be expert problem solvers to contribute meaningfully. The result will be a faster rate of progress—solving more problems, faster and better. And, as history has shown, the supply of open research problems is virtually unbounded. For example, in the context of AI systems alone, workloads are becoming more complex (e.g., test-time compute, online reinforcement learning), hardware is growing more heterogeneous (e.g., specialized accelerators, new networking technologies), and performance and scalability demands are higher than ever (e.g., multi-modal training and inference).

\section{Conclusion}
\label{sec:conclusion}

As AI systems take over algorithm discovery, the role of the human researcher will change. Much like an academic advisor guides a student, the researcher of the future will act as a guide for these AI systems. The researchers' responsibilities will continue to be defining the problem, steering the research process, and critically evaluating the results.

One of the most profound implication is the potential for a \emph{virtuous cycle}. We can use an \SYS/ to improve itself. As recent work has shown, models can learn to refine their own reasoning, debug their code, and discover more effective strategies. As these AI agents rapidly improve themselves, they will compound the pace of scientific discovery.

To summarize, in this paper, we have illustrated the potential of \SYS/ in systems research. Our case studies show that the \SYS/ approach can already outperform human baselines on key performance problems. While still very early, we call on the systems community to embrace these tools, not just as accelerators of research, but as subjects of research. Improving \SYS/--making them efficient, scalable, and reliable--is itself a systems challenge. As such, we believe that system builders are uniquely positioned to shape the future of AI-driven discovery.
\section*{Acknowledgments}
We thank Aurojit Panda, Tianyin Xu, Asankhaya Sharma, Rishabh Iyer, and Dave Patterson for their valuable feedback and insightful discussions. 
This research was supported by gifts from Accenture, AMD, Anyscale, Broadcom Inc., Google, IBM, Intel, Intesa Sanpaolo, Lambda, Mibura Inc, Samsung SDS, and SAP.

\bibliography{iclr2025_conference}
\bibliographystyle{iclr2025_conference}


\newpage
\appendix
\section{Ablation Studies}
\label{appendix-1}
We have conducted a few ablation studies on the Can't Be Late case study and provide the results as follows. 

\subsection{GEPA results} In addition to OpenEvolve, we apply GEPA to evolve the program generation for the Can't Be Late use case. We build a custom adapter that wraps two stages of evaluation, similar to OpenEvolve: stage 1 validates syntax and simulator compliance, while stage 2 runs full simulations on our test workloads to compute cost savings. 
GEPA uses this score as the fitness value and attaches diagnostic feedback for failed runs.
Feedback is then converted into a reflective dataset which guides a model (we use OpenAI o3 in this setup) to rewrite the candidate program. 
We initialize the process with the greedy baseline, and set our evaluation metric as the average cost savings across single-region traces. 

We give the greedy program as the base program, and use Claude Opus to refine the base program entirely. The feedback function given to GEPA is the score evaluated on the downstream single-region traces. Our final result achieved 4\% improvement over the greedy baseline at iteration 68, capping the search at 200 iterations.




\subsection{Changing Train Set Coverage}
\begin{figure}[h!]
  \centering
  \begin{subfigure}[t]{0.48\textwidth}
    \centering
    \includegraphics[width=\textwidth]{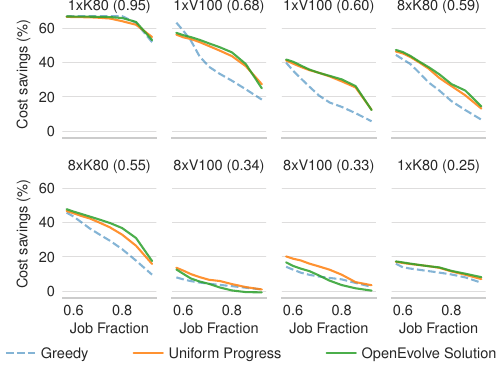}
    \caption{Training with 3\% of the available training set.}
    \label{fig:failure-taxonomy-a}
  \end{subfigure}
  \hfill
  \begin{subfigure}[t]{0.48\textwidth}
    \centering
    \includegraphics[width=\textwidth]{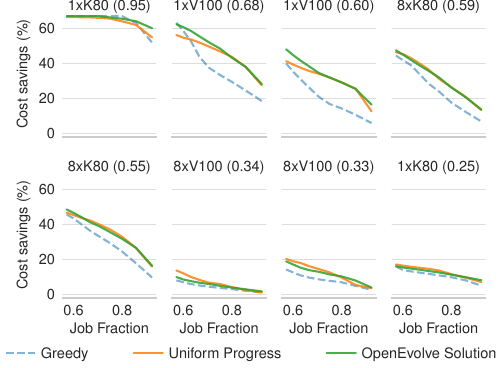}
    \caption{Training with the full training set.}
    \label{fig:failure-taxonomy-b}
  \end{subfigure}
  \caption{Impact of training set coverage on policy evolution. 
  We split data into 30\% training and 70\% testing. 
  The left panel shows results when using only 3\% of the training data, 
  while the right panel uses the entire training set.}
  \label{fig:failure-taxonomy}
\end{figure}




\newpage
\section{Failure Taxonomy}
\label{sec:failure-taxonomy}
ADRSes demonstrate exciting capabilities in advancing systems research in our case study; however, they currently face challenges that require careful setup. To mitigate these challenges, we first characterize the common failure modes ADRSes exhibit. In the next section (see Section~\ref{sec:best-practices}), we outline best practices and guidelines to address these limitations.


As shown in Table~\ref{tab:f}, we group failures into three broad categories—execution errors, search (evolution) failures, and algorithm failures, and report their frequency in our evaluation. Detailed descriptions of each failure type are provided in Section~\ref{sec:best-practices}.

\textbf{Execution Errors.} Among one third of the failures come from immediate execution failures. 
The most common cases are \textit{syntax and interface errors}, where the generated solution code fails to compile or connect with the evaluator due to missing imports, type mismatches, or misusing required interfaces. 
Other fails due to \textit{budget exhaustion}, where a candidate consumes excessive memory, runs into timeouts, or exceeds model API quotas. 

\textbf{Search Failures.} Roughly half of the failiures happen when the solution executes but search fails to make progress.
\textit{Premature convergence} happens when the search fo solution settles on a local optimum too early, such as sticking with a steiner tree strategy in multi-region transfer while ignoring opportunities to make use of data partitions. 
\textit{Stuck-in-the-loop} refers to generating near-duplicate solutions without improvement, like iterations that only rename variables or add logging. 
\textit{Mutation drift}occurs when edits are contradictory or random. For example, oscillating between BFS tree and graph-based method in the multi-region network transfer task, slowing or preventing convergence.

\textbf{Algorithm Failures.} These are failures where candidate solutions run but fail to advance algorithms. 
Common patterns include \textit{misaligned objectives}, where the solution ignore constraints (e.g., throughput gains that break latency SLOs). 
\textit{Suboptimal optimizations} also appear frequently as the evolved solution makes shallow tweaks in APIs instead of coming up with algorithmic innovation.
\textit{Overfitting} hard-codes behavior to evaluation traces (e.g., spot scheduling policies collapsing on unseen spot patterns).
\textit{Reward hacking} exploits loopholes in the evaluator. For example, in our multi-agent runs, some candidates bypass a required stage in algorithm, causing the evaluator to assign higher scores.

\section{Case Studies}
We now present detailed evolution results from the case studies detailed in Table~\ref{tab:project-summary} as follows. 


\subsection{Adaptive weight compression} This task studies adaptive weight quantization with variable per-column bitrate. The objective is to assign high vs. low bitrates across columns to minimize overall storage cost (measured in average bits per element across all elements in the weight tensor, lower is better) while preserving accuracy (measured by Wikitext-2 perplexity, or PPL, where lower indicates higher accuracy). The base program is a hand-written heuristic that applies dynamic bitrate encoding, where each column of the weight matrix is quantized at a bitrate chosen according to its importance score. An initial hand-tuned calculation and mapping from importance score to column bitrate provided reasonable compression performance of 2.64 bits/elem and PPL of 22.9 on a Llama3.2-1B, but left room for improvement.  

To guide evolution, the evaluator combines normalized scores from Wikitext PPL (0.5), PTB PPL (0.2), and bitrate (0.3), with solutions exceeding 2.5 bits/elem and PPl of 30 receiving zero score, to hit a compression ratio under 2.5 with the highest accuracy. We use Gemini-2.5-pro to generate candidates, running 200 iterations in about 12 hours with less than 20 USD. The evolution takes more than 10 hours due to the evaluator time, i.e., each evaluation needs to fully quantize the entire Llama3.2-1B model before computing perplexity and bitrate. The evolved program achieves 2.49 bits/elem with PPL of 22.84, hitting a lower bitrate compared to the best sweep configuration (2.5 bits/elem) while maintaining accuracy. 

The improved solution from the evolution simply tunes hyperparameter values of the hand-crafted mapping between importance scores and bitrates, without any algorithmic improvements on column importance finding or mapping strategy. As such, it offers only marginal gains over a parameter sweep and remains more of an automated hyperparameter tuner than a novel quantization strategy. Future work could explore structural changes to better prompting (e.g., explicitly listing the methods and parameters to explore) or evolutionary settings (e.g., exploration ratio) to move beyond incremental improvements.

\subsection{Network Telemetry Repair}
This task identified by HotNets’24~\cite{krentsel2024case} studies repair of faulty router telemetry signals, which can become buggy due to router faults or collection infrastructure errors. Such inconsistencies (for example, counters on the two ends of a link not matching) can cause the network controllers to make incorrect decisions. The objective is to detect and repair faulty telemetry to produce a self-consistent view of network state.  

We use OpenEvolve to evolve repair strategies, running 300 iterations (about 8 hours) with a mix of reasoning (GPT-o3, 80\%) and non-reasoning (GPT-4o, 20\%) models, supplemented by contextual hints from the HotNets’24 paper. The evolved program introduces structured repair logic, including averaging nearby counters to reduce noise, and separating repair and confidence estimation into distinct steps. It achieves a repair score of 95\% and confidence calibration of 95\%, outperforming the HotNets’24 solution (86\% repair, 65\% confidence).

\subsection{Cloudcast}
The Cloudcast problem, published in NSDI ’24~\cite{wooders2024cloudcast}, studies cost-aware multicast across multi-region and multi-cloud topologies. The objective is to construct an overlay connecting a source, multiple destinations, and optional waypoints so as to minimize egress cost.  

Our initial program is a direct replication strategy: the source sends data independently to each destination. While simple, this approach often incurs high egress costs when destinations are located in distant or expensive regions. To guide evolution, the evaluator tests candidate algorithms across 10 multi-region, multi-cloud configurations and assigns a total score based on the overall egress cost of the scheduled paths. We use OpenEvolve with a mix of o3 and Gemini-2.5-pro models, running 100 iterations in about one hour, costing less than \$10 for the entire run.  

The evolved solution successfully rediscovers the Steiner tree strategy as the state-of-the-art, achieving an average cost reduction of 31.1\% compared to direct replication. This solution constructs a cost-efficient multicast tree by introducing intermediate waypoints. For example, data may be replicated once at a cheaper waypoint region and then forwarded to multiple destinations, reducing the total egress cost.

\subsection{Global model placement} The research problem~\cite{yu2025prism} focus on the challenge of multi-LLM serving on shared GPUs under bursty, heterogeneous workloads. The key metric is the KV pressure ratio (KVPR), defined as the SLO-weighted request rate divided by available KV cache memory. The optimization goal is to minimize the maximum KVPR across GPUs, thereby reducing contention and improving SLO compliance.  

The base program is the algorithm from the paper~\cite{yu2025prism}: models are placed sequentially onto the first GPU with sufficient remaining memory, without considering long-term balance. While feasible, this approach often overloads some GPUs while leaving others underutilized.  
We use OpenEvolve to evolve improved placement strategies, guided by a scoring function that combines execution correctness with the KVPR objective. The evolution runs with GPT-4o (70\%) and GPT-o3 (30\%) as the model ensemble, converging in about 70 iterations (roughly 40 minutes). The best evolved program achieves an 18.5\% higher score compared to the state-of-the-art reported in the original paper.  

The evolved strategy mirrors the SOTA algorithm from the paper and assignment used in PRISM, but crucially adds a local improvement stage. After the initial placement, it repeatedly tests whether moving or swapping models between GPUs reduces the maximum KVPR, applies such refinements, and stops once it finds a move that can reduce the current maximum KVPR. 

\accheng{TBU to sparse attention}
\subsection{Sparse attention design} The research problem~\cite{hashattention} studies efficient sparse attention mechanisms that approximate dense attention while balancing accuracy and compute efficiency. 
Sparse attention reduces computation by restricting each query to attend to only a subset of keys, but selecting this subset (i.e., the active indices) introduces a trade-off between density (fraction of active indices) and relative error (approximation quality). 
The objective is therefore to evolve sparse attention masks that minimize a combined loss of density and relative error, formulated as $\text{Score} = -(\text{Density} + \text{Relative Error})$. 


\subsection{Multi-Agent System Optimization} 
The research problem extends the work of~\cite{cemri2025multi} (NeurIPS'25), which studies diagnosing and repairing failures in multi-agent LLM systems (MAS), and proposes MAST as the taxonomy of MAS failure modes. Such systems (e.g., MetaGPT~\cite{metagpt}, ChatDev~\cite{chatdev}) often suffer from breakdowns in coordination, memory, or communication that reduce task success. The challenge is how to improve multi-agent systems, evolving more robust agent architectures, prompts, and inter-agent communication patterns to increase reliability and performance.  

The base program is a direct adaptation of MetaGPT~\cite{metagpt}, assembled into a minimal Python implementation (roughly 400 LOC). This initial version defines fixed agent roles, communication protocols, and system prompts, but frequently encounters the failure modes identified in the MAST. To guide evolution, the evaluator uses the MAST annotator, which assigns a score of $1/(1+\text{total FM occurrence})$, penalizing common coordination and memory failures.  

We ran three evolution configurations with different mutation scopes: (1) modifying agent definitions and communication schemes, (2) introducing new verification and communication flows using GPT-5, and (3) allowing changes to the number and types of agents, i.e., the system topology. The evolved solutions introduced innovations such as improved context management (v1) and verification/communication flows (v2), though removing verification in v3 degraded performance. Overall, downstream program development success improved from 40\% in the base program to 47\% (v1) and 53\% (v2) on the ProgramDev-v1 benchmark, before dropping to 30\% in v3. The fast that verification agent was removed in v3 was an example of reward hacking (since we penalize the verification failures, the evolution algorithm got rid of the whole verification when it could) and an example of the importance of carefully tuning which parts of the initial code the evolution algorithm is allowed to change. These results show that OpenEvolve can automatically discover MAS design refinements that improve robustness, though careful control over mutation scope is critical to avoid reward hacking.


\subsection{GPU Cache Algorithm Optimization} 
This task studies optimizing cache algorithms for GPU execution. While CPU versions of the algorithm run correctly, directly adapting them to GPUs often introduces significant overhead and poor performance.  

The base program was an initial GPU implementation generated by GPT-o3. This version was extremely slow and failed some correctness tests, making it unsuitable for practical use. To guide evolution, the evaluator balanced two objectives: correctness (passing functional tests) and runtime speed.  

We ran 120 iterations of OpenEvolve using Gemini-2.5-flash, with a total runtime of about 40 minutes and cost of roughly \$20. The evolved program uncovered practical optimizations such as wrapping key kernels with \texttt{torch.jit}, which noticeably improved execution speed. However, despite these gains, the final code remained slower than desired and continued to fail certain edge-case tests.  


\end{document}